\documentclass[pdflatex,sn-basic,Numbered,iicol]{sn-jnl}

\usepackage{graphicx}%
\usepackage{multirow}%
\usepackage{amsmath,amssymb,amsfonts}%
\usepackage{amsthm}%
\usepackage{mathrsfs}%
\usepackage[title]{appendix}%
\usepackage{xcolor}%
\usepackage{textcomp}%
\usepackage{manyfoot}%
\usepackage{booktabs}%
\usepackage{algorithm}%
\usepackage{algorithmicx}%
\usepackage{algpseudocode}%
\usepackage{listings}%
\usepackage{pifont}
\newcommand{\cmark}{\ding{51}}
\newcommand{\xmark}{\ding{55}}
\usepackage{makecell} 
\usepackage{dblfloatfix} 
\usepackage{etoolbox}
\usepackage{float}

\makeatletter

\AtBeginDocument{%
  \@ifundefined{bibfont}
    {}
    {}%
  \@ifundefined{bibsep}
    {}
    {\setlength{\bibsep}{0pt plus 0.1ex}}%
}

\patchcmd{\keywords}
  {\par\addvspace{10pt}}
  {\par\addvspace{4pt}}
  {}
  {\PackageWarning{main}{Could not patch \string\keywords}}

\patchcmd{\@maketitle}
  {\removelastskip\vskip36pt\vskip0pt}
  {\removelastskip\vskip8pt\vskip0pt}
  {}
  {\PackageWarning{main}{Could not patch \string\@maketitle}}

\makeatother

\theoremstyle{thmstyleone}%
\theoremstyle{thmstyletwo}%

\theoremstyle{thmstylethree}%

\begin{document}

\title[Panoramic Scene Understanding: A Survey]{Panoramic Scene Understanding: A Survey from Distortion-Aware Engineering to Sphere-Native Modeling}








\author[1,2]{\fnm{Qinfeng} \sur{Zhu}}

\author*[2]{\fnm{Lei} \sur{Fan}}\email{lei.fan@xjtlu.edu.cn}

\affil[1]{\orgname{University of Liverpool}, \orgaddress{\country{United Kingdom}}}

\affil[2]{\orgname{Xi'an Jiaotong-Liverpool University}, \orgaddress{\country{China}}}


\abstract{Panoramic images capture the complete visual sphere in a single frame, providing spatial context unavailable to conventional cameras. Yet this completeness comes at a geometric cost: the 2-sphere cannot be faithfully mapped to the plane, and every planar representation introduces distortions that violate the assumptions underlying standard vision architectures. This survey traces the evolution of panoramic scene understanding from projection-based adaptation, through distortion-aware engineering, to sphere-native modeling, a trajectory of deepening geometric commitment to the sphere. Foundation models add a distinct fourth family, geometry-aware tokenization, that adapts the input interface while reusing perspective-pretrained weights. We evaluate these approaches across five task families: dense prediction, unified multi-task understanding, open-world perception, vision-language reasoning, and dynamic video analysis. Our review reveals that the same drift toward spherical geometry recurs within each task. In practice, however, the field has converged not on the theoretically strongest sphere-native operators, which attain exact rotation equivariance but cannot reuse perspective-pretrained backbones and so have never scaled, but on a compatibility-preserving middle ground that couples moderate geometric awareness with large pretrained models. This geometric commitment is markedly uneven across tasks: it runs deepest in dense prediction and shallowest in dynamic perception, where methods stay spatially sphere-aware yet temporally planar, and foundation-model adaptation has advanced panoramic depth fastest while leaving layout, surface-normal, and video-level understanding largely untouched. To date, no panoramic foundation model has been pretrained directly on spherical data. We further identify five gaps in current evaluation protocols, namely the absence of spherical-area-weighted metrics, seam-consistency testing, polar-robustness stratification, cross-projection generalization, and open-world protocol standardization, and distill our analysis into a six-point roadmap toward general-purpose panoramic intelligence. The corresponding repository is publicly available at: \url{https://github.com/zhuqinfeng1999/Awesome-Panoramic-Scene-Analysis}.}




\maketitle


\section{Introduction}
\label{sec:introduction}

Panoramic images capture the full $360^{\circ} \times 180^{\circ}$ visual sphere, encoding the complete spatial context of a scene in a single frame. This property makes them useful for applications that require holistic environmental awareness, including virtual and augmented reality, autonomous navigation, robotic perception, and immersive media~\cite{Gao2022TIM}. With the commoditization of consumer $360^{\circ}$ cameras and the growing need for spatial intelligence in embodied AI~\cite{Zheng2025PANORAMA}, panoramic scene understanding has attracted increasing attention from the computer vision community.

A panoramic image, however, is not simply a large field-of-view (FoV) perspective image. The underlying signal lives on the 2-sphere $\mathbb{S}^2$, a compact Riemannian manifold with constant positive curvature whose isometry group is $\mathrm{SO}(3)$ (rotations) rather than $\mathbb{R}^2$ (translations). Any projection to a planar grid therefore introduces distortions: the ubiquitous equirectangular projection (ERP) produces area distortion that diverges toward the poles and a boundary discontinuity at the seam. These are not minor perturbations. They conflict with the translation-invariant receptive-field assumption that underlies convolutional architectures~\cite{Coors2018SphereNet}, break the spatial uniformity expected by vision transformers, and cause systematic failure when perspective-pretrained models are applied to panoramic inputs. Zhang~et~al.~\cite{Zhang2021DensePASS} benchmarked more than 20 representative pinhole-trained segmenters and reported a roughly $50\%$ average mIoU drop when these models were evaluated on panoramic data without adaptation, with both CNN- and transformer-based architectures showing comparable orders of degradation across their tested set.

In response, the panoramic vision community has developed four broad families of strategies, each pushing geometric structure deeper into the model:

\begin{enumerate}
\item \textbf{Projection-based adaptation} decomposes the panorama into perspective patches, such as tangent images on an icosahedron~\cite{Eder2020Tangent} or cubemap faces, and applies unmodified perspective models, or fuses multi-projection features as in BiFuse~\cite{Wang2020BiFuse}. This avoids architectural changes at the cost of boundary discontinuity and redundant computation.

\item \textbf{Distortion-aware engineering} operates directly on ERP but modifies convolution kernels~\cite{Coors2018SphereNet, Tateno2018Distortion}, attention mechanisms~\cite{Zhang2022Trans4PASS, Shen2022PanoFormer}, or padding strategies to compensate for the analytic distortion profile.

\item \textbf{Sphere-native modeling} abandons planar representations and defines operators on $\mathbb{S}^2$ or $\mathrm{SO}(3)$ via spherical harmonics~\cite{Cohen2018Spherical, Esteves2018SO3}, icosahedral meshes~\cite{Cohen2019Gauge}, or graph convolutions on equal-area grids~\cite{Defferrard2020DeepSphere}, obtaining rotation equivariance at the cost of incompatibility with standard 2D operators and pretrained planar models.

\item \textbf{Foundation-model interface adaptation} (also referred to as \emph{geometry-aware tokenization}, abbreviated GT in the rest of this paper) redesigns the input interface to steer large pretrained models toward panoramic competence while keeping their weights compatible with perspective pretraining. Examples include geometry-aware positional encoding (ERP-RoPE in Dense360~\cite{Dense360_2025}), spherical patch embedding (SGAT4PASS~\cite{Li2023SGAT4PASS}), patch-sequence reformulation for SAM2 memory (OmniSAM~\cite{Zhong2025OmniSAM}), and trajectory-based perspective scanning (SAP~\cite{Jiang2026SAP}).
\end{enumerate}

We organize these four families along a methodological trajectory of increasing geometric commitment, from treating distortion as a nuisance to be corrected to treating the sphere as the native domain of the data. The trajectory is conceptual rather than strictly chronological: sphere-native methods such as S$^2$CNN~\cite{Cohen2018Spherical} and gauge-equivariant networks~\cite{Cohen2019Gauge} appeared as early as 2018--2019, well before some of the distortion-aware Transformers and foundation-model adapters discussed below, but they have so far failed to gain traction at scale because of their incompatibility with pretrained planar backbones. A largely orthogonal axis concerns training and transfer: the paradigm has shifted from supervised learning on scarce panoramic data, through unsupervised domain adaptation (UDA)~\cite{Ma2021DensePASS, Zhang2022Trans4PASS, Zheng2024_360SFUDAplus}, to foundation-model-assisted transfer~\cite{Zhang2024GoodSAM, Zhong2025OmniSAM}.

Several surveys have previously addressed panoramic vision. Gao~et~al.~\cite{Gao2022TIM} reviewed panoramic imaging systems and their optical underpinnings, with a focus on hardware and multidimensional sensing. Ai~et~al.~\cite{Ai2022DeepOmni} surveyed deep learning for omnidirectional vision with broad coverage of tasks and representations. More recently, Lin~et~al.~\cite{Lin2025FlightSurvey} presented a cross-task survey covering more than 300 papers and over 20 representative tasks, with particular emphasis on the perspective-to-panorama adaptation gap. Zheng~et~al.~\cite{Zheng2025PANORAMA} proposed a system-level architecture for omnidirectional vision in embodied AI.

While our survey shares part of its scope with these works, it differs in organizing principle and extends coverage to recent topics they treat only briefly. Whereas Lin~et~al.\ organize panoramic vision horizontally across more than 20 tasks and over 300 papers, we organize it \emph{vertically} along a methodological axis. Three choices distinguish our treatment. First, we adopt a two-dimensional taxonomy (architecture $\times$ paradigm) that exposes the trade-off structure of the design space rather than listing methods task by task. Second, we cover the 2025--2026 frontier in detail, including vision-language understanding on panoramas, open-world perception, and temporal panoramic analysis; this material lies largely outside the earlier surveys of Gao~et~al.~\cite{Gao2022TIM} and Ai~et~al.~\cite{Ai2022DeepOmni} and is only partially covered by the concurrent 2025 surveys~\cite{Lin2025FlightSurvey, Zheng2025PANORAMA}. Third, we critically examine evaluation protocols and identify five systematic gaps that current benchmarks fail to probe.

Specifically, this survey makes the following contributions:

\begin{itemize}
\item \textbf{A geometry-first formulation and taxonomy.} We formalize panoramic scene understanding as inference on $\mathbb{S}^2$ and organize the literature with a two-dimensional taxonomy (architectural design $\times$ training paradigm) that traces a trajectory of increasing geometric commitment and exposes the trade-off structure of the design space, including its unexplored regions (Sections~\ref{sec:geometry}--\ref{sec:ch3}).

\item \textbf{A unified, task-spanning treatment.} Rather than cataloguing methods task by task, we follow a single geometric thread through dense prediction, unified multi-task understanding, open-world perception, vision-language reasoning, and dynamic video analysis (Sections~\ref{sec:scene_understanding}--\ref{sec:ch6}), and argue that spatial completeness does not by itself imply spatial intelligence.

\item \textbf{A critique of evaluation and a research agenda.} We expose systematic gaps in current evaluation protocols and in the surveyed tasks that keep reported numbers from reflecting genuine spherical understanding, and distill them into a concrete roadmap toward general-purpose panoramic intelligence (Sections~\ref{sec:datasets}--\ref{sec:open_problems}).
\end{itemize}

The remainder of this paper is organized as follows. Section~\ref{sec:geometry} establishes the geometric foundations and problem formulation. Section~\ref{sec:ch3} presents the methodological taxonomy. Section~\ref{sec:scene_understanding} surveys panoramic scene understanding from closed-set to open-world. Section~\ref{sec:ch5} examines vision-language understanding and spatial reasoning. Section~\ref{sec:ch6} covers dynamic panoramic perception. Section~\ref{sec:datasets} systematizes datasets and evaluation protocols. Section~\ref{sec:open_problems} discusses open problems and future directions. Section~\ref{sec:conclusion} concludes the survey.
 
\section{Panoramic Imaging Geometry and Problem Formulation}
\label{sec:geometry}

This section establishes the geometric foundations that distinguish panoramic scene understanding from perspective vision. We trace the imaging pipeline from capture through spherical geometry to planar representations, identify the challenges that arise, and formulate panoramic understanding as inference on the 2-sphere $\mathbb{S}^2$, a problem governed by rotational rather than translational symmetry.

\subsection{Panoramic Imaging Pipeline}
\label{sec:imaging_pipeline}

\begin{figure*}[tp]
\centering
\includegraphics[width=0.93\textwidth]{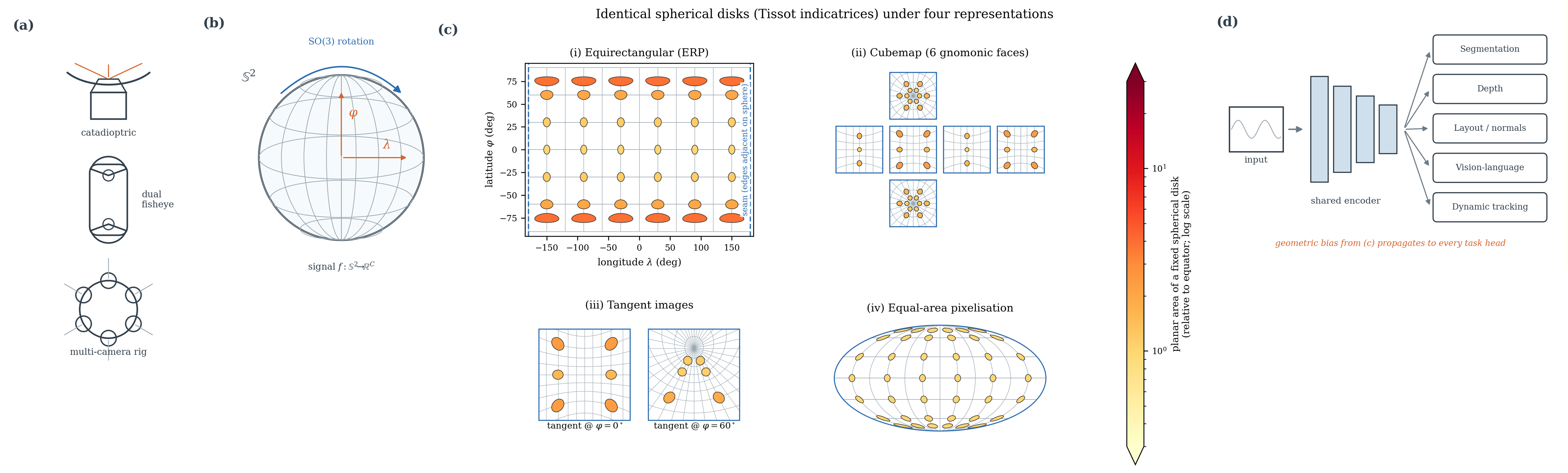}
\caption{Panoramic imaging pipeline. Diverse capture systems (a) produce signals on the unit sphere (b), which must be projected to planar representations (c) before entering learning pipelines (d). The projection choice determines which geometric biases propagate downstream.}
\label{fig:pipeline}
\end{figure*}

Rather than reproduce a full optical-system taxonomy (see Gao~et~al.~\cite{Gao2022TIM} for a comprehensive treatment), we focus on the four pipeline stages of Fig.~\ref{fig:pipeline} that determine what a learning algorithm receives as input. Panoramic images originate from physically diverse sensors (catadioptric mirrors, fisheye lenses, multi-camera rigs; Fig.~\ref{fig:pipeline}a), yet all converge to the same mathematical abstraction: Geyer and Daniilidis~\cite{Geyer2000Unifying} proved that every central panoramic system is equivalent to a projection onto the unit sphere $\mathbb{S}^2$ (Fig.~\ref{fig:pipeline}b) via $\mathbf{x}_s = \mathbf{X}/\|\mathbf{X}\|$, followed by a parameterized mapping to an image plane. Subsequent models, e.g. Kannala--Brandt~\cite{Kannala2006Generic}, the Enhanced UCM~\cite{Khomutenko2015EUCM}, and the Double Sphere model~\cite{Usenko2018DoubleSphere}, refine this second step, but the unit sphere remains the canonical intermediate representation. Polydioptric configurations (e.g., dual-fisheye consumer cameras) require stitching, which introduces residual parallax and chromatic artifacts~\cite{DaSilveira2022Survey360}.

Because current deep learning infrastructures require rectangular arrays, the spherical signal $f:\mathbb{S}^2\to\mathbb{R}^C$ must be projected onto planar grids (Fig.~\ref{fig:pipeline}c) before processing. This projection step is where representation-specific biases enter the pipeline: every subsequent convolution, pooling, and loss computation, and ultimately every task head (Fig.~\ref{fig:pipeline}d), inherits the geometric properties and distortions of the selected mapping.

\subsection{Projection Spaces and Representation Choices}
\label{sec:projections}

No isometric mapping exists between the sphere and the plane. Gauss's \emph{Theorema Egregium} states that Gaussian curvature is invariant under local isometries; since $\mathbb{S}^2$ has constant Gaussian curvature $K=1/R^2>0$ while any open region of the Euclidean plane has $K=0$, no isometric map from a region of $\mathbb{S}^2$ to the plane can exist. Because a smooth map that simultaneously preserves area and local angles is necessarily an isometry, no such map exists either: every planar parameterization must distort areas, angles, or both. Topology adds a separate constraint: $\mathbb{S}^2$ is a closed manifold without boundary, so any global mapping to a rectangular planar domain must either cut the sphere along a one-dimensional curve (introducing a seam) or collapse a measure-zero set such as the poles to a degenerate strip; the equirectangular projection used throughout this survey does both. Fig.~\ref{fig:pipeline}(c) makes these effects concrete by showing how one fixed spherical disk deforms under each representation family, and Table~\ref{tab:projection_comparison} summarizes their trade-offs.

\paragraph{Equirectangular projection.}
ERP maps longitude $\lambda$ and latitude $\varphi$ directly to image coordinates ($x\!=\!\lambda,\; y\!=\!\varphi$), producing a $W\!\times\!H$ image with $W\!=\!2H$ (Fig.~\ref{fig:pipeline}c-i). Its simplicity makes it the dominant storage and processing format. However, the solid angle per pixel is $\mathrm{d}\Omega = \cos\varphi\;\mathrm{d}\lambda\;\mathrm{d}\varphi$, so pixel area distortion grows as $1/\cos\varphi$, diverging at the poles~\cite{Zhang2022Trans4PASS}. A fixed convolutional kernel therefore covers vastly different solid angles depending on its latitude, violating the translation-invariant receptive-field assumption~\cite{Coors2018SphereNet, Stringhini2024SWHDC}.

\paragraph{Perspective-patch representations.}
Both cubemap (Fig.~\ref{fig:pipeline}c-ii) and tangent-image (Fig.~\ref{fig:pipeline}c-iii) representations project the sphere onto local gnomonic (perspective) planes. Gnomonic projection maps every great circle to a straight line and preserves angles exactly only at the tangent point; it is neither conformal nor equal-area, but the angular distortion within a single tangent patch (and \emph{a fortiori} within icosahedral subdivisions) is far smaller than that of ERP near the poles. Inter-patch discontinuity remains the dominant cost. The cubemap uses six axis-aligned $90^{\circ}$ faces; cube padding~\cite{Cheng2018CubePad} and spherical padding~\cite{Wang2020BiFuse} partially address boundary artifacts, though object integrity remains orientation-dependent~\cite{Xiong2018SnapAngle}. Tangent images~\cite{Eder2020Tangent} generalize this to icosahedral subdivisions with lower per-patch distortion but greater overlap redundancy.

\paragraph{Non-planar representations.}
HEALPix~\cite{Gorski2005HEALPix} provides equal-area pixelization on iso-latitude rings (Fig.~\ref{fig:pipeline}c-iv), enabling graph-based spherical CNNs such as DeepSphere~\cite{Defferrard2020DeepSphere} with approximate rotation equivariance. Spherical harmonics (SH) offer rotation equivariance via spectral convolution~\cite{Cohen2018Spherical, Esteves2018SO3} that is exact in the continuous, bandlimited formulation; in any finite numerical implementation it holds only up to discretization and quadrature error, and the ``exact'' label used in Tables~\ref{tab:projection_comparison} and~\ref{tab:ch3_comparison} should be read in this sense. The cost is $O(B^3)$ in the bandwidth $B$ and a loss of spatial locality. Both families sacrifice compatibility with standard 2D convolution libraries, limiting practical adoption.

No representation dominates across all criteria (Table~\ref{tab:projection_comparison}), which reflects the underlying topology of $\mathbb{S}^2$.

\begin{table}[htbp]
\centering
\footnotesize
\setlength{\tabcolsep}{4pt}
\caption{Panoramic representation trade-offs. \cmark: satisfied; \xmark: violated; $\sim$: partial. SH: spherical harmonics.}
\label{tab:projection_comparison}
\begin{tabular}{@{}p{1.9cm}ccccc@{}}
\toprule
\textbf{Property} & \textbf{ERP} & \textbf{Cubemap} & \textbf{Tangent} & \textbf{HEALPix} & \textbf{SH} \\
\midrule
Area preservation         & \xmark & \xmark & \xmark & \cmark & N/A \\
Local angle preservation  & \xmark & $\sim^\dagger$ & $\sim^\dagger$ & $\sim$ & N/A \\
Topological continuity    & $\sim^\ddagger$ & \xmark & \xmark & \cmark & \cmark \\
Rectangular grid          & \cmark & \cmark & \cmark & \xmark & \xmark \\
Rotation equivariance     & \xmark & \xmark & \xmark & $\sim$ & \cmark \\
Pretrained model reuse    & \cmark & \cmark & \cmark & \xmark & \xmark \\
\bottomrule
\end{tabular}
\par\smallskip
{\raggedright\scriptsize $^\dagger$Gnomonic projection is non-conformal; angle preservation is exact only at each face's tangent point and degrades toward the face boundary. $^\ddagger$Periodic in longitude; singular at poles.\par}
\end{table}

\subsection{Fundamental Geometric Challenges of Panoramic Representations}
\label{sec:challenges}

Several challenges in panoramic scene understanding are structural rather than superficial. ERP's non-uniform sampling means that a fixed-pixel kernel near latitude $80^{\circ}$ subtends only $\cos 80^{\circ} \approx 0.174$ of the equatorial solid angle, i.e.\ roughly $5.76$ times less spherical area than the same kernel at the equator, which distorts spatial operations built on regularity assumptions~\cite{Yu2023OSRT, Stringhini2024SWHDC}. Boundary discontinuity compounds this: ERP's left and right edges ($\lambda\!=\!\pm\pi$) are physically adjacent on the sphere but treated as unrelated by standard zero-padding. Islam~et~al.~\cite{Islam2020Padding} showed in the general planar setting that zero-padding implicitly encodes absolute position, an inductive bias that is appropriate for perspective images, where absolute position is informative, but spurious on the rotationally symmetric domain of a panorama. This cost has since been quantified directly in the panoramic regime: Jiang~et~al.~\cite{Jiang2025DA360} reported measurable depth-error increases at ERP boundaries that disappear once circular padding replaces zero padding, and Wu~et~al.~\cite{Wu2026_360Anything} traced seam artifacts in panoramic generation to VAE encoder padding rather than to the generative process itself. Object shape variance adds a further complication: the same physical object at different ERP latitudes is horizontally stretched by $1/\cos\varphi$, so latitude-agnostic features lose accuracy near the poles~\cite{Coors2018SphereNet}.

We distinguish two categories of \emph{geometric} challenges that matter for method design, and they are not specific to any single projection. \emph{Projection-induced artifacts}, including polar stretching, boundary discontinuity, and metric bias, arise from the chosen representation and can be partially alleviated by modifying or combining projections. \emph{Intrinsic spherical properties}, including the absence of global translation symmetry (since $\mathbb{S}^2$ admits only rotations as isometries), path-dependent parallel transport, and the impossibility of distortion-free flattening, are independent of representation and require architectures that respect the geometry of the sphere. This distinction corresponds to the methodological landscape outlined in Section~\ref{sec:ch3}: distortion-aware methods address the former, while sphere-native methods address the latter.

\subsection{Why Perspective Priors Do Not Transfer Directly}
\label{sec:transfer_failure}

Beyond the headline mIoU drops summarized in Section~\ref{sec:introduction}, the perspective-to-panoramic gap is structural rather than parametric. The DensePASS benchmark~\cite{Zhang2021DensePASS} showed that the degradation is consistent across more than 20 architectures, and that transformer-based models such as SETR~\cite{Zheng2021SETR} are not immune. At the foundation model scale, Cao~et~al.~\cite{Cao2025PanDA} showed that Depth Anything models are sensitive to spherical spatial transformations that are trivial on $\mathbb{S}^2$, and Zhang~et~al.~\cite{Zhang2026MTPano} revealed that rotation-invariant tasks (depth, segmentation) and rotation-variant tasks (surface normals) cause mutual negative transfer on panoramic data, a conflict absent in the perspective domain.

Generic fine-tuning without geometry-aware adaptation is insufficient to close these failures. At the data level, real labeled panoramic segmentation benchmarks contain at most a few thousand annotated panoramas (e.g., DensePASS 100, WildPASS 500, Stanford2D3D 1{,}413), and even a densely multi-task-annotated synthetic panoramic corpus such as Structured3D~\cite{Zheng2020Structured3D}, with $\sim$21K rendered rooms, is several orders of magnitude smaller than the perspective pretraining corpora that drive modern foundation models. At the architecture level, translation equivariance is structural, encoded in the weight-sharing pattern of planar convolution, and cannot be replaced by rotation equivariance through weight adjustment; Kondor and Trivedi~\cite{Kondor2018Equivariance} proved that convolutional structure is both sufficient and necessary for equivariance to compact group actions. This requirement extends beyond the convolution operator: even icosphere-based spherical Transformers can collapse under full $\mathrm{SO}(3)$ rotation when their positional encodings capture gravity-aligned latitude cues rather than intrinsic geometry (Section~\ref{sec:sphere_native}), so geometric fidelity must be maintained at every architectural level. At the representation level, a distortion-compensating weight tuned for one location on the sphere is necessarily wrong at another, because the distortion is position-dependent (in ERP, for example, compensation calibrated at one latitude fails at others). The perspective-to-panoramic gap is therefore better understood as a geometry mismatch than as a conventional domain shift, one that additional perspective-style fine-tuning cannot remove.

\subsection{scene understanding on the Sphere: A Formal Problem Statement}
\label{sec:formulation}

Sections~\ref{sec:imaging_pipeline}--\ref{sec:transfer_failure} established that central or stitched panoramic systems are commonly modeled as spherical signals, that no projection of the 2-sphere preserves them faithfully, that the sphere's geometric properties differ fundamentally from the plane, and that perspective architectures fail systematically on this domain. Building on this, we state panoramic scene understanding directly on its native domain, following the standard equivariance formulation of geometric deep learning~\cite{Cohen2018Spherical, Kondor2018Equivariance}. Let $f:\mathbb{S}^2\to\mathbb{R}^C$ be a panoramic observation and $\mathcal{T}$ a task mapping $f$ to an output $g:\mathbb{S}^2\to\mathbb{R}^K$. The fundamental requirement on a model $\Phi$ implementing $\mathcal{T}$ is equivariance to the domain symmetries:
\begin{equation}
  \Phi(L_R \cdot f) = \rho(R) \cdot \Phi(f), \quad \forall\, R \in G,
  \label{eq:equivariance}
\end{equation}
where $G$ is the symmetry group (in principle $\mathrm{SO}(3)$; in practice often reduced to azimuthal $\mathrm{SO}(2)$ by gravity), $L_R$ acts on inputs, and $\rho(R)$ on outputs.

This formulation imposes three design requirements on practical systems, and these requirements structure our taxonomy in Section~\ref{sec:ch3}. The first is \emph{distortion awareness}: most systems operate on ERP or cubemap, so they need mechanisms that compensate for or are invariant to projection-specific distortions. The second is \emph{sphere-native operations}: tasks demanding geometric fidelity require operators defined directly on $\mathbb{S}^2$ with appropriate equivariance, such as spherical convolution~\cite{Cohen2018Spherical}, gauge-equivariant layers~\cite{Cohen2019Gauge}, or graph convolutions on spherical grids~\cite{Defferrard2020DeepSphere}. The third is \emph{pretrained-model compatibility}: the knowledge encoded in perspective foundation models (CLIP, DINOv2, Depth Anything, SAM) must be accessible to panoramic methods without inheriting the geometric biases documented above.

Section~\ref{sec:ch3} organizes existing approaches into a taxonomy built around these three requirements, tracing an evolutionary trajectory from distortion-aware engineering to sphere-native modeling.

\section{Learning on the Sphere: A Taxonomy of Panoramic Modeling}
\label{sec:ch3}

\begin{figure}[t]
\centering
\includegraphics[width=0.93\linewidth]{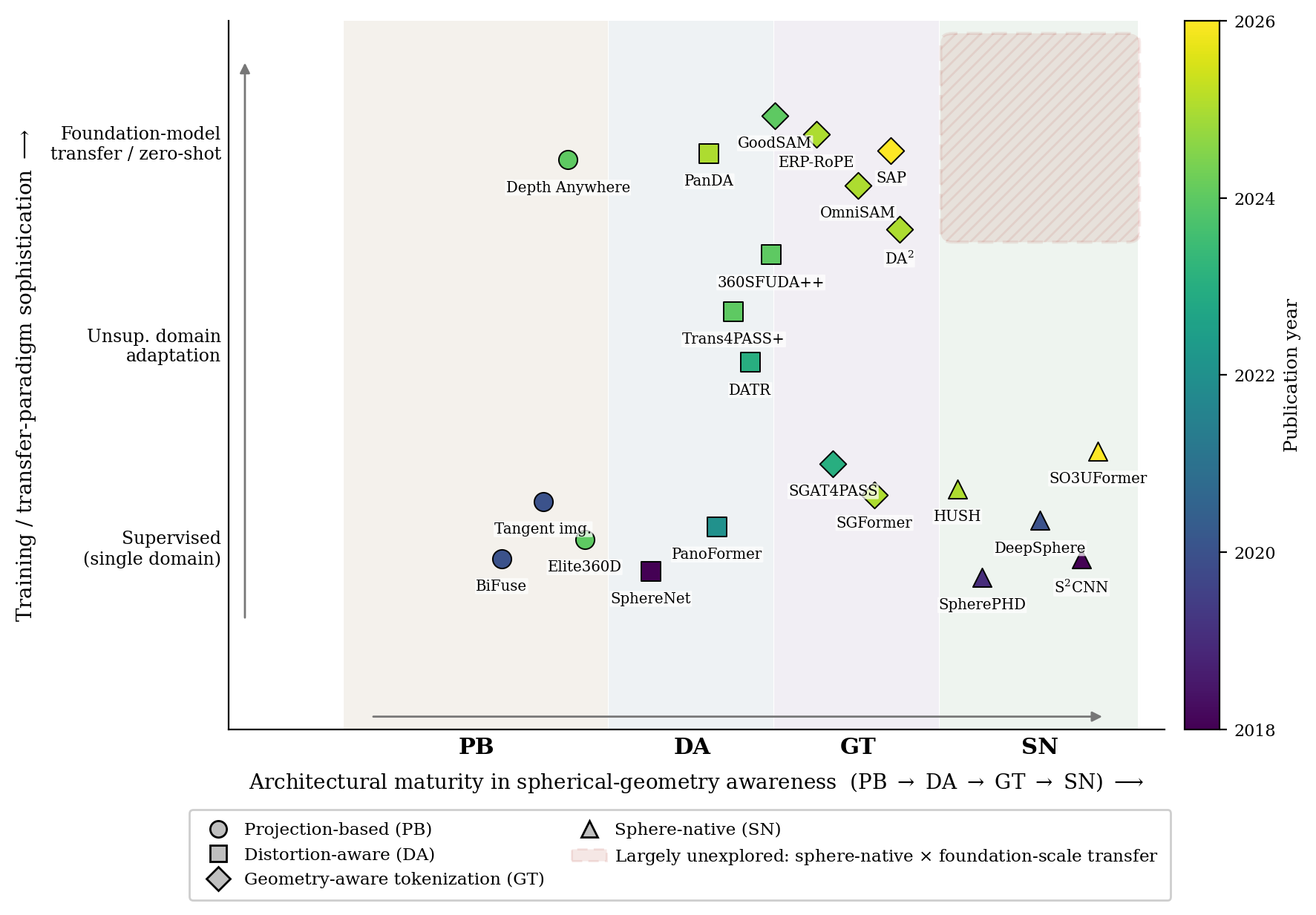}
\caption{Two-dimensional taxonomy of panoramic modeling. The horizontal axis
represents architectural sophistication in spherical geometric awareness; the
vertical axis represents training/transfer paradigm complexity. Methods are
color-coded by era. The upper-right region (sphere-native $\times$ foundation-model
transfer) remains largely unexplored.}
\label{fig:ch3_taxonomy}
\end{figure}

This section develops the core methodological taxonomy for panoramic modeling,
organized along two orthogonal dimensions (Fig.~\ref{fig:ch3_taxonomy}).
\emph{Dimension~A}, architectural and representational design
(Sections~\ref{sec:proj_adapt} to \ref{sec:geo_token}), traces the progression from
adaptation based on projection, through distortion-aware engineering, to
modeling defined directly on the sphere and geometry-aware tokenization for foundation models.
\emph{Dimension~B}, training and transfer paradigms (Section~\ref{sec:domain_adapt}),
examines how knowledge is transferred across domains, moving from perspective imagery to panoramic imagery,
from synthetic data to real data, and from foundation models to task-specific networks.
We organize the taxonomy along two axes because, historically, architectural awareness of geometry and sophistication of training paradigms advanced largely independently: early methods typically innovated on one axis while taking the other as fixed. Only recently have the strongest methods begun to combine innovations along both axes, which is precisely why a two-axis view is needed to place them.
Table~\ref{tab:ch3_overview} provides a roadmap of the method families discussed in
this section, and Table~\ref{tab:ch3_comparison} presents a comprehensive comparison
across eight evaluation axes.

\begin{table*}[tp]
\caption{Roadmap of panoramic modeling families discussed in this section. The first four entries are architectural families, mutually distinguished by how operators interact with spherical geometry; domain adaptation is presented as an orthogonal transfer paradigm that combines with any of them.}
\label{tab:ch3_overview}
\begin{minipage}{\textwidth}
\centering
\begin{tabular}{@{}p{0.11\textwidth}p{0.14\textwidth}p{0.21\textwidth}p{0.43\textwidth}@{}}
\toprule
\textbf{Section} & \textbf{Axis} & \textbf{Family / Paradigm} & \textbf{Representative Works} \\
\midrule
\ref{sec:proj_adapt} & Architecture & Projection-based & Tangent~\cite{Eder2020Tangent}, BiFuse~\cite{Wang2020BiFuse}, CViT~\cite{bai2024glpanodepth} \\
\ref{sec:distort_aware} & Architecture & Distortion-aware & SphereNet~\cite{Coors2018SphereNet}, Trans4PASS~\cite{Zhang2022Trans4PASS} \\
\ref{sec:sphere_native} & Architecture & Sphere-native & S$^2$CNN~\cite{Cohen2018Spherical}, SpherePHD~\cite{Lee2019SpherePHD} \\
\ref{sec:geo_token} & Architecture & FM interface adapt.\ (GT) & ERP-RoPE~\cite{Dense360_2025}, SGAT4PASS~\cite{Li2023SGAT4PASS}, OmniSAM~\cite{Zhong2025OmniSAM}, SAP~\cite{Jiang2026SAP} \\
\ref{sec:domain_adapt} & \emph{Transfer} & Domain adapt.\ (orthogonal) & DensePASS~\cite{Ma2021DensePASS}, 360SFUDA++~\cite{Zheng2024_360SFUDAplus} \\
\bottomrule
\end{tabular}

\smallskip
\footnotesize HUSH~\cite{Lee2025HUSH} sits between SN and GT (Section~\ref{sec:sphere_native}); we treat it as SN by default.
\end{minipage}
\end{table*}

Throughout the rest of this survey we refer to these four architectural families (Table~\ref{tab:ch3_overview}) by the acronyms introduced here: \emph{PB} for projection-based, \emph{DA} for distortion-aware, \emph{SN} for sphere-native, and \emph{GT} for geometry-aware tokenization (throughout this survey \emph{GT} denotes geometry-aware tokenization, not ground truth). Foundation-model adaptation, abbreviated \emph{FMA}, is conceptually a transfer paradigm rather than an architectural family: an FMA method takes any of the architectures above and trains it under foundation-model supervision (pseudo-labels, LoRA fine-tuning, distillation). We therefore treat FMA as a modifier and use compound labels for methods that combine an architectural family with FMA-style training (e.g., DA$+$FMA for a distortion-aware backbone supervised by a foundation-model teacher, GT$+$FMA for a tokenization-level method that fine-tunes a pretrained foundation backbone).

\subsection{Projection-Based Adaptation via Planar Decomposition}
\label{sec:proj_adapt}

The most direct strategy for applying deep networks to panoramic images is to
convert the spherical signal into planar views compatible with standard
architectures. Tangent-image methods render the sphere onto locally undistorted
patches on a subdivided icosahedron, enabling direct application of pretrained
CNNs without architectural modification~\cite{Eder2020Tangent}; this approach has
been adopted for depth estimation via multi-patch fusion with geometric feature
compensation~\cite{Li2022OmniFusion, ReyArea2022_360MonoDepth}. Cubemap-based
methods partition the sphere into six perspective faces and address boundary
discontinuities through cube padding~\cite{Cheng2018CubePad} or spherical
padding~\cite{Wang2020BiFuse}; more recently, the Cubemap Vision Transformer (CViT) introduced in GLPanoDepth~\cite{bai2024glpanodepth} extends this paradigm to ViT backbones, leveraging global receptive fields on each distortion-free face to provide coherent predictions for spherical signals. Rather than committing to a single projection, fusion
methods combine complementary representations: BiFuse~\cite{Wang2020BiFuse}
introduced bidirectional ERP--cubemap fusion, UniFuse~\cite{Jiang2021UniFuse}
simplified this to efficient unidirectional fusion at the decoding stage, and
Elite360M~\cite{Ai2024Elite360M} further extended fusion to multi-task learning
with icosahedron-based sampling points. The progression from bidirectional to unidirectional fusion suggests that symmetric feature exchange may not repay its cost, though no controlled comparison under matched backbones and budgets yet exists. What these methods share is that distortion is sidestepped by locality rather than modeled: each planar patch is only approximately undistorted, and the residual artifacts plus inter-patch seams motivate the distortion-aware approaches of the next subsection.

\subsection{Distortion-Aware Operators and Architectures}
\label{sec:distort_aware}

A second family of methods operates directly on equirectangular projection (ERP)
but engineers explicit distortion compensation into the network. Compared to
projection-based methods, these approaches avoid the overhead of multi-branch
processing and boundary fusion, but they require careful geometric modeling of
the projection's distortion profile. Three principal convolutional strategies
address the spatially varying distortion of ERP, each representing a different
point in the trade-off between geometric precision, computational cost, and
pretrained model compatibility.
Su and Grauman~\cite{Su2017Pano} proposed row-wise untied kernel weights learned
through cross-projection distillation. SphereNet~\cite{Coors2018SphereNet} and
Tateno et al.~\cite{Tateno2018Distortion} independently developed geometry-driven
sampling grid deformation, adjusting filter sampling locations via tangent-plane
projection to ``wrap'' kernels around the sphere. Notably, Tateno et al.'s
formulation enables training on perspective images and deployment on panoramas
without panoramic training data. The Kernel Transformer Network
(KTN)~\cite{Su2019KTN} learns a compact function that transforms source
perspective kernels into latitude-dependent spherical kernels, offering model
transferability across different source architectures. These three strategies
span a spectrum from purely geometric (SphereNet) to purely learned (KTN)
distortion compensation, and all operate within the standard ERP representation
and remain tied to its distortion profile. They are complemented by cube padding~\cite{Cheng2018CubePad} and spherical
padding~\cite{Wang2020BiFuse} that connect adjacent cubemap face boundaries, by
circular padding along the longitudinal seam for ERP-based networks, and by
latitude-aware loss weighting that rescales pixel contributions by their solid angle $\cos\varphi$ to prevent over-emphasizing polar
regions~\cite{Li2023SGAT4PASS}.


The Vision Transformer offers several distinct points at which a distortion prior can be injected, and the methods in this family are best understood by which point they target rather than as a flat list. The earliest inject it at \emph{tokenization}: Trans4PASS~\cite{Zhang2022Trans4PASS} learns latitude-dependent offsets through Deformable Patch Embedding (DPE) and Deformable MLP (DMLP) at both the patch-extraction and feed-forward stages, paired with Mutual Prototypical Adaptation (MPA) for unsupervised domain adaptation; its TPAMI extension Trans4PASS+~\cite{Zhang2024Trans4PASS} upgrades DMLP to a parallel-token-mixing DMLPv2, strengthens MPA with pseudo-label rectification, and contributes the SynPASS dataset for synthetic-to-real adaptation. A second group injects the prior at \emph{token sampling}: PanoFormer~\cite{Shen2022PanoFormer} forms tokens on the spherical tangent domain and uses a learnable token flow that selects relevant tokens per query, achieving distortion-aware attention without explicit geometric formulas. A third injects it at the \emph{attention bias}: EGFormer~\cite{Yun2023EGFormer} turns the known ERP distortion profile into a bias on vertical and horizontal window attention, reaching strong depth accuracy at the lowest cost among transformer-based methods. A fourth restricts attention \emph{locality}: DATR~\cite{Zheng2023DATR} builds a backbone that captures the pixel-wise neighbor correlations specific to ERP distortion and couples it with class-wise feature aggregation, showing that distortion-aware architecture and transfer paradigm can be co-designed. Across all four injection points, the prior is tied to one projection's distortion profile: a method built around ERP cannot readily transfer to a different projection, whereas a method that models the sphere itself would in principle be projection-agnostic. This shared limitation motivates the sphere-native approaches of the next subsection.

\subsection{Sphere-Native Modeling}
\label{sec:sphere_native}

Sphere-native methods abandon planar representations entirely and define
computational primitives directly on the sphere or its discrete approximations,
offering theoretical guarantees such as exact rotation equivariance at the cost of
higher computation and incompatibility with pretrained planar models. In
practice, adoption of these methods has been limited by the computational gap: a
spectral spherical convolution at bandwidth $B=128$ (roughly equivalent to a
$256\times256$ ERP image) already requires orders of magnitude more computation
than a standard planar convolution at similar resolution.
The theoretical foundations were laid by Cohen et al.~\cite{Cohen2018Spherical},
who defined an exactly rotation-equivariant $\mathbb{S}^2$ cross-correlation whose output
lives on $\mathrm{SO}(3)$ and computed it via a generalized (non-commutative) Fast Fourier Transform (FFT),
and by Esteves et al.~\cite{Esteves2018SO3}, who performed convolutions directly
in the spherical harmonic domain with filters whose structure is equivalent to
the zonal case on the sphere, reducing both parameter count and per-layer cost
and keeping intermediate features on $\mathbb{S}^2$ rather than $\mathrm{SO}(3)$.
Subsequent analyses show that, at bandwidth $B$, spectral spherical CNNs remain
asymptotically more expensive than planar convolutions at comparable resolution,
which has limited their adoption at scale~\cite{Ocampo2022DISCO,Esteves2023Scaling}. Subsequent work extended expressivity via spin-weighted spherical
functions~\cite{Esteves2020Spin} and scaled these architectures to practical
problems through hardware-optimized implementations~\cite{Esteves2023Scaling}.

Alternative discrete approaches improve scalability but, in exchange, achieve only approximate rather than exact rotation
equivariance. SpherePHD~\cite{Lee2019SpherePHD} represents omnidirectional images
on a subdivided icosahedron with near-uniform spatial resolving power, enabling
standard CNN operations on the polyhedron mesh. Cohen et al.~\cite{Cohen2019Gauge}
generalized equivariance to local gauge transformations on the icosahedron, while
Zhang et al.~\cite{Zhang2019OrientAware} argued that orientation awareness (not
rotation invariance) is preferable for semantic segmentation with known camera
orientation, aligning hexagonal filters to the north pole to preserve gravity-based
semantic cues. This design represents an opposite stance to the equivariance
pursued by Cohen et al., trading symmetry exploitation for task-specific prior
knowledge. This trade-off, however, can fail silently: a model that leans on gravity-aligned orientation rather than true equivariance breaks once that orientation assumption is violated. This failure mode is exactly the one previewed in Section~\ref{sec:transfer_failure}, and the reported numbers below make it concrete. On Stanford2D3D segmentation, the SphereUFormer baseline~\cite{benny2025sphereuformer} reportedly drops from $67.53\%$ to $25.26\%$ mIoU when evaluated under arbitrary $\mathrm{SO}(3)$ rotations of the input panorama, while keeping its training-time gravity alignment fixed~\cite{zhu2026so3uformer}. The same study reports that SO3UFormer~\cite{zhu2026so3uformer}, which combines gauge-aware relative positional encoding in local tangent planes, quadrature-consistent attention correcting for non-uniform icosphere sampling, and a logit-level $\mathrm{SO}(3)$-consistency regularizer, retains $70.67\%$ mIoU under the same stress test. Independent reproduction of these stress-test numbers is not yet available; we therefore treat them as illustrative of the gravity-lock failure mode rather than as a settled benchmark result. Graph-based approaches, notably
DeepSphere~\cite{Perraudin2019DeepSphere, Defferrard2020DeepSphere}, represent
the sphere as a HEALPix graph and apply Chebyshev graph convolutions with
controllable equivariance error, achieving performance comparable to spectral
methods at lower cost, which suggests that anisotropic filters may be
unnecessary for many tasks. At the representation level,
HUSH~\cite{Lee2025HUSH} treats spherical-harmonic bases as a geometric inductive bias rather than a strictly equivariant operator, sitting between the sphere-native methods of this section and the GT methods of Section~\ref{sec:geo_token}; the full hierarchical-attention mechanism is detailed in Section~\ref{sec:unified}.

On the question of whether sphere-native methods are truly necessary, Gerken et al.~\cite{Gerken2022EquivAug} provide the most controlled evidence to date, although the experiments use spherical MNIST and spherical FashionMNIST rather than real-world panoramic scenes. They find that for spherical image classification, sufficient data augmentation and parameter scaling allow non-equivariant CNNs to match equivariant S$^2$CNNs, whereas for spherical semantic segmentation the non-equivariant networks saturate well below their equivariant counterparts even with heavy augmentation. Whether this gap carries over to real-world panoramic dense prediction at modern dataset and model scale remains an open empirical question, since no comparable controlled study has been reported on Stanford2D3D or Matterport3D. The persistent challenge that holds independently of this question is incompatibility with pretrained planar models, which motivates the approaches in the next subsection.

\subsection{Foundation-Model Interface Adaptation}
\label{sec:geo_token}

The emergence of large pretrained models (ViT, SAM/SAM2, multimodal large language models / MLLMs) creates a new
challenge: how to feed spherical data into architectures with deeply ingrained planar
assumptions without sacrificing pretrained knowledge. Unlike the methods in
Sections~\ref{sec:distort_aware} to \ref{sec:sphere_native}, which redesign the model
core for spherical computation, the methods grouped here keep the backbone close to its
perspective-pretrained form and engineer the \emph{interface} between spherical inputs
and planar foundations.We distinguish five sub-strategies by where the spherical prior is injected: (i) input-stage positional encoding and patch embedding (ERP-RoPE~\cite{Dense360_2025}, SDPE~\cite{Li2023SGAT4PASS}); (ii) decoder-side spherical priors (SGFormer~\cite{Zhang2025SGFormer}); (iii) sequence re-construction aligned with SAM2's video memory (OmniSAM~\cite{Zhong2025OmniSAM}, SAP~\cite{Jiang2026SAP}); (iv) distillation from perspective foundation models (GoodSAM~\cite{Zhang2024GoodSAM}, GoodSAM++~\cite{Zhang2024GoodSAMplus}); and (v) kernel-level re-instantiation onto the sphere without retraining~\cite{Liu2025SphSegCompat}. The unifying property is that the backbone weights remain compatible with their perspective-pretrained initialization. Their deployment in specific vision-language model (VLM) systems is deferred to Section~\ref{sec:ch5}.

The first two strategies keep a single backbone and inject the prior at its input or its decoder. Standard 2D RoPE~\cite{Heo2024RoPEViT} assumes a planar grid, which overlooks
two geometric facts about ERP: the horizontal axis wraps around at the date
line, and the pixel-to-angle ratio varies with latitude. Several recent works
close this gap from different angles.
Dense360~\cite{Dense360_2025} proposes \emph{ERP-RoPE}, an extension of
multimodal RoPE that handles horizontal continuity along latitude circles
and compensates for the latitude-dependent drop in information density
toward the poles. It is a drop-in positional encoding that requires no
change to the backbone architecture.
SGFormer~\cite{Zhang2025SGFormer} injects spherical priors on the decoder
side instead. Its Spherical Prior Decoder combines bipolar re-projection,
circular rotation, and curve local embedding to preserve equidistortion,
continuity, and surface distance respectively, and uses a query-based Global
Conditional Position Embedding (GCPE) to reintroduce spatial structure
across resolutions.
SGAT4PASS~\cite{Li2023SGAT4PASS} moves the spherical prior to the patch
embedding stage: its Spherical Deformable Patch Embedding (SDPE) augments
deformable patch embedding with intra- and inter-patch geometric constraints
derived from the spherical surface, paired with a panorama-aware loss that
accounts for non-uniform pixel density.

The remaining three strategies reuse an unmodified foundation model by reframing its input, distilling from a teacher, or re-instantiating its weights on the sphere. GoodSAM~\cite{Zhang2024GoodSAM} and GoodSAM++~\cite{Zhang2024GoodSAMplus} take a teacher--student route, with task-level evaluation deferred to Section~\ref{sec:foundation}.
OmniSAM~\cite{Zhong2025OmniSAM} (compound label \emph{GT$+$UDA}: see Section~\ref{sec:domain_adapt} for its UDA formulation) repurposes SAM2's video memory mechanism by treating sliding-window panoramic patches as video frames, letting the memory attention extract cross-patch correspondences that stand in for spatial continuity. The task-level evaluation, including its UDA performance on indoor and outdoor benchmarks, is deferred to Section~\ref{sec:foundation}.
SAP~\cite{Jiang2026SAP} pushes this idea further by decomposing each panorama
into overlapping perspective views sampled along a continuous spherical
trajectory, which better satisfies the smooth-motion assumption behind
SAM2's memory propagation, and by synthesizing 183K 4K panoramas with
instance masks for large-scale fine-tuning.
Liu et al.~\cite{Liu2025SphSegCompat} take a different route: they re-sample
the kernel locations of existing planar convolutions onto the sphere,
turning a pretrained CNN into a sphere-compatible network without
retraining its weights. This sits at the intersection of the
distortion-aware paradigm (Section~\ref{sec:distort_aware}) and the
sphere-native paradigm (Section~\ref{sec:sphere_native}).

None of these methods fundamentally redesigns the core architecture of the
foundation model for spherical data. They engineer the interface between
spherical inputs and planar models, and the development of a truly sphere-native
foundation model remains a central open question
(Section~\ref{sec:open_problems}).

\subsection{Domain Adaptation and Transfer Paradigms}
\label{sec:domain_adapt}

The training/transfer dimension is largely orthogonal to architectural
design: any architecture from
Sections~\ref{sec:proj_adapt}--\ref{sec:geo_token} can be combined with the
transfer strategies discussed here. The need for transfer arises from a
persistent data asymmetry. Densely annotated perspective images abound
(e.g., Cityscapes~\cite{Cordts2016Cityscapes} with 5,000 fine-annotated
images), while panoramic annotations remain scarce and expensive because of
the larger spatial extent and the annotation difficulty induced by
distortion.

Domain adaptation for Pin2Pan has progressively moved from generic alignment toward signals that exploit panoramic geometry. The setting was formalized in the DensePASS line: Ma~et~al.~\cite{Ma2021DensePASS} introduced the densely annotated benchmark and a basic UDA setup, and Zhang~et~al.~\cite{Zhang2021DensePASS} extended it into the P2PDA framework, aligning the domains in the output, feature, and feature-confidence spaces with uncertainty-aware attention. The Trans4PASS / Trans4PASS+ line (architecture in Section~\ref{sec:distort_aware}) replaced this with prototype-level alignment through Mutual Prototypical Adaptation, and Trans4PASS+~\cite{Zhang2024Trans4PASS} added pseudo-label rectification and synthetic-to-real transfer via SynPASS. The decisive step toward geometry-specific adaptation is DPPASS~\cite{Zheng2023DPPASS}, which decomposes the Pin2Pan gap into an \emph{inherent} gap (style and scene differences) and a \emph{format} gap (projection differences) and handles each separately: a dual-path framework processes ERP and tangent images in parallel, combining tangent-wise contrastive learning with cross-projection prediction consistency, while intra-projection training adds adversarial alignment. Crucially, the tangent path can be dropped at inference, so cost matches a single-branch baseline. By turning the geometric correspondence between two projections of the same sphere into a free supervisory signal, DPPASS exploits structure that has no analog in planar-to-planar adaptation; we refer to this signal, which recurs in several later methods, as \emph{multi-projection self-supervision}.

More restrictive settings have also been explored.
360SFUDA++~\cite{Zheng2024_360SFUDAplus} targets source-free adaptation,
where only a pretrained source model (not the source data) is available,
and it pushes multi-projection self-supervision further: tangent
projection and fixed-FoV projection extract knowledge from the source
model, while a reliable panoramic prototype adaptation module combined
with a cross-projection dual attention module transfers this knowledge to
the target panoramic domain, improving over prior SFUDA baselines across
synthetic and real-world benchmarks.
DTA4PASS~\cite{Jiang2025DTA4PASS} defines multi-source adaptation,
jointly exploiting real pinhole and synthetic panoramic images. Its
Unpaired Semantic Morphing module turns pinhole images into
geometrically distorted ones through an adversarial diffeomorphic
deformation network trained under a dual-view discriminator, and a
Distortion Gating Alignment module performs feature-level matching.

Foundation-model-assisted transfer covers a spectrum.
At one end, GoodSAM~\cite{Zhang2024GoodSAM} and its extension
GoodSAM++~\cite{Zhang2024GoodSAMplus} learn a compact panoramic student
without any labeled panoramic data, by combining SAM's class-agnostic
instance masks with a perspective-pretrained teacher assistant that
supplies semantic labels and distilling the ensemble through several dedicated adaptation modules.
At the other end, OmniSAM~\cite{Zhong2025OmniSAM}, whose architectural side was discussed in Section~\ref{sec:geo_token}, is simultaneously a fully specified UDA method on the training/transfer axis: it fine-tunes the SAM2 image encoder with LoRA on labeled source data (e.g., SPin8 or CS13) and adapts to unlabeled target panoramas via FoV-based prototypical adaptation and dynamic pseudo-label updating. Quantitative results on Pin2Pan benchmarks are reported in Section~\ref{sec:foundation}.
The shared property of these methods is that the foundation model
serves as a transferable visual prior; the practical labeling
requirement, however, ranges from zero panoramic and zero source labels
(GoodSAM family) down to standard Pin2Pan UDA (OmniSAM).

\subsection{Unifying Comparison and Synthesis}
\label{sec:unify_compare}

Table~\ref{tab:ch3_comparison} synthesizes the method families of
Sections~\ref{sec:proj_adapt}--\ref{sec:domain_adapt} across eight
dimensions. Four patterns stand out.

First, geometric fidelity and pretrained compatibility form the
central trade-off. Spectral spherical CNNs achieve exact equivariance
but cannot load ImageNet-pretrained weights, while geometry-aware PE
methods keep full compatibility at the cost of only approximate
geometric awareness tied to a specific projection. In practice,
benchmarks have converged on the middle ground: moderate geometric
awareness combined with strong pretrained compatibility. The $\mathrm{SO}(3)$ collapse reported in Section~\ref{sec:sphere_native}
shows that fidelity must be carried through positional encoding and
attention normalization, not the representation alone.

Second, \emph{multi-projection self-supervision} (introduced as
cross-projection consistency in Section~\ref{sec:domain_adapt}) is a
signal unique to panoramic data. Unlike planar domain adaptation, where
source and target are two different views of the world, panoramic
adaptation can exploit the geometric correspondence between ERP,
tangent, and cubemap projections of the \emph{same} spherical signal as
free supervision~\cite{Zheng2023DPPASS, Zheng2024_360SFUDAplus}. No
analog exists in planar-to-planar adaptation, and the gains from this
signal have been reproduced across several UDA frameworks.

Third, current panoramic foundation models are sophisticated
projection pipelines rather than architectural redesigns. ERP-RoPE
modifies only the positional encoding; OmniSAM reframes panoramas as
video-like patch sequences for SAM2's memory; SAP constructs perspective
scanning trajectories that better fit SAM2's temporal assumptions;
GoodSAM distills SAM and a semantic teacher assistant into a compact
student through auxiliary modules. None of them rewires the backbone
for spherical input: the core representation remains perspective-native,
and all the spherical-awareness lives at the input/output interface.

Fourth, the two-dimensional design space still has an empty quadrant. In Fig.~\ref{fig:ch3_taxonomy} this is the upper-right region, where the horizontal axis (sphere-native architecture) meets the top of the vertical axis (foundation-model-scale training); it remains largely unexplored. Most of the strong recent methods already combine moves along
both axes of the design space. Trans4PASS+ pairs a distortion-aware
backbone with prototypical adaptation; OmniSAM pairs foundation-model
adaptation with domain-adaptive fine-tuning. Whether the empty quadrant
can be reached will depend on co-designing architecture and training
paradigm around the specific geometry of the sphere
(Section~\ref{sec:open_problems}).

\begin{table*}[tp]
\centering
\caption{Unified comparison of panoramic modeling families across eight dimensions. \textbf{Sph.~Compat.}: spherical geometry compatibility (none / approx.\ / exact). \textbf{Proj.~Dep.}: projection dependence. \textbf{Training Signal}: dominant supervisory regime under which the family is trainable (perspective-transferable supervision, panoramic data only, or task-specific). \textbf{FM~Trans.}: foundation-model transferability. \textbf{X-Proj.}: cross-projection generalization. TP: tangent projection.}
\label{tab:ch3_comparison}
\resizebox{\textwidth}{!}{%
\begin{tabular}{l|c|c|c|c|c|c|c|c}
\hline
\textbf{Method Family} & \textbf{Representation} & \textbf{Operator} & \textbf{Sph.\ Compat.} & \textbf{Proj.\ Dep.} & \textbf{Cost} & \textbf{Training Signal} & \textbf{FM Trans.} & \textbf{X-Proj.} \\
\hline\hline
Tangent/cubemap (\S\ref{sec:proj_adapt}) & TP / cube faces & Std.\ conv/ViT & Approx. & Proj.-specific & Med. & Persp.\ transf. & Direct & Limited \\
Multi-proj.\ fusion (\S\ref{sec:proj_adapt}) & ERP + cube/TP & Conv + fusion & Approx. & Multi-proj. & Med--High & Persp.\ transf. & Partial & Moderate \\
Dist.-aware conv (\S\ref{sec:distort_aware}) & ERP & Deform.\ conv & Approx. & ERP-specific & Light & Persp.\ transf. & Re-arch. & Limited \\
Dist.-aware Transf.\ (\S\ref{sec:distort_aware}) & ERP & Deform.\ PE/attn & Approx. & ERP-specific & Med. & Partial & Partial & Limited \\
Spectral sph.\ CNN (\S\ref{sec:sphere_native}) & $\mathbb{S}^2$/SO(3) & SH conv & \textbf{Exact} & \textbf{Agnostic} & High & Sph.\ data & Incompat. & \textbf{Native} \\
Ico./graph mesh (\S\ref{sec:sphere_native}) & Mesh / graph & Mesh/graph conv & Approx. & Agnostic & Med. & Limited & Re-arch. & Approx. \\
Geometry-aware PE (\S\ref{sec:geo_token}) & ERP & ViT + geo-PE & Approx. & ERP-specific & \textbf{Light} & Persp.\ transf. & \textbf{High} & Limited \\
FM interface adapt.\ (\S\ref{sec:geo_token}) & ERP$\to$TP/seq.\ & SAM/ViT $+$ adapt. & Approx.\ (interface) & ERP / persp.-view & Med--High & \makecell[c]{Varies:\\label-free / UDA /\\synthetic-supervised} & \textbf{Direct} & Limited \\
\hline
Domain adapt.\ (\S\ref{sec:domain_adapt}) & \multicolumn{8}{c}{\textit{Orthogonal to architecture: combinable with any family above}} \\
\hline
\end{tabular}%
}
\end{table*}

\section{Panoramic Scene Understanding: From Closed-Set to Open-World}
\label{sec:scene_understanding}

This section traces how panoramic \emph{dense prediction}, the family of pixelwise tasks in which a model assigns a label to every pixel rather than a single label to the whole image, has evolved from closed-set supervised learning, through unified multi-task representations and foundation-model adaptation, toward open-world perception. We focus on how spherical geometry reshapes each task, namely semantic segmentation, depth estimation, room layout estimation, and their unification, rather than re-explaining the architectural operators cataloged in Section~\ref{sec:ch3}. Tasks whose outputs are language-grounded belong to Section~\ref{sec:ch5}; temporal and video-level perception is covered in Section~\ref{sec:ch6}.

\subsection{Closed-Set Dense Prediction}
\label{sec:closed_dense}

\paragraph{Panoramic segmentation.}
Panoramic semantic segmentation (PASS) is the task where the geometric obstacles of Section~\ref{sec:geometry} were first felt, and its development reads as a sequence of responses to them. The starting problem is that perspective-trained networks fail on unfolded panoramas; the earliest PB methods, the original PASS~\cite{Yang2019PASS,Yang2020PASS} and the WildPASS-era ECANet~\cite{Yang2021WildPASS}, addressed this by adapting perspective networks to the panorama, with ECANet adding concurrent horizontal and vertical attention to recover the omni-range context lost by a narrow receptive field. Once ERP distortion itself became the bottleneck, the distortion-aware Transformer wave led by Trans4PASS~\cite{Zhang2022Trans4PASS} and its TPAMI successor Trans4PASS+~\cite{Zhang2024Trans4PASS} embedded deformable patch embeddings and deformable MLPs into the pipeline and, through mutual prototypical adaptation, made unsupervised domain adaptation the dominant way to cope with scarce panoramic labels. Subsequent work targeted the costs this paradigm exposed: DATR~\cite{Zheng2023DATR} cut parameters by roughly 80\% (4.64M vs.\ 24.98M) while improving Synthetic-to-Real mIoU by over 8\%, by restricting attention to the less-distorted local neighborhood, and SGAT4PASS~\cite{Li2023SGAT4PASS} exposed a different weakness, showing through its Spherical Geometry-Aware (SGA) validation that prior methods such as Trans4PASS+ lose an order of magnitude in stability under small pitch/roll perturbations. Most recently, the open problem of compute cost has motivated three Mamba-based architectures, Deformable Mamba~\cite{Hu2024DeformableMamba}, Mamba4PASS~\cite{xu2025mamba4pass}, and MCCL4PASS~\cite{xu2025mamba}, which reach competitive accuracy far more cheaply, though whether they will decisively surpass Transformers remains open. The training-paradigm axis is summarized in Section~\ref{sec:domain_adapt}, and foundation-model-assisted adaptation via GoodSAM~\cite{Zhang2024GoodSAM} and OmniSAM~\cite{Zhong2025OmniSAM} in Section~\ref{sec:foundation}. Two task variants extend the setting beyond flat semantics: panoramic panoptic segmentation, introduced with the WildPPS dataset by Jaus~et~al.~\cite{Jaus2023PanopticPano}, and occlusion-aware seamless segmentation at the ERP seam, addressed by OASS~\cite{Cao2024OASS}.

\paragraph{Panoramic depth estimation.}
Monocular $360^{\circ}$ depth estimation has developed along four technical routes, summarized in Table~\ref{tab:pano_depth}. The \emph{bi-projection fusion} route, whose architectural taxonomy was introduced in Section~\ref{sec:proj_adapt}, has progressed within depth from BiFuse~\cite{Wang2020BiFuse} and UniFuse~\cite{Jiang2021UniFuse} (ERP$+$cubemap, bidirectional then unidirectional), through HRDFuse~\cite{Ai2023HRDFuse} (ERP$+$tangent, holistic-regional collaboration), to Elite360D~\cite{Ai2024Elite360D}, whose substitution of cubemap faces by an icosahedron projection (ICOSAP) achieves substantially more uniform solid angle per face and reaches state-of-the-art accuracy through a Bi-projection Bi-attention Fusion (B2F) module that captures semantic- and distance-aware dependencies between ERP pixels and ICOSAP points. The \emph{Transformer} route models the long-range dependencies needed for wide-FoV depth: PanoFormer~\cite{Shen2022PanoFormer} tokenized spherical tangent-domain patches with learnable token flow; EGFormer~\cite{Yun2023EGFormer} and SGFormer~\cite{Zhang2025SGFormer} incorporated equirectangular geometry biases and spherical position embeddings respectively. The \emph{tangent decomposition} route, exemplified by 360MonoDepth~\cite{ReyArea2022_360MonoDepth} and OmniFusion~\cite{Li2022OmniFusion}, sharply reduces ERP distortion within each gnomonic patch but sacrifices global context, leaves residual gnomonic warp inside large patches, and incurs expensive cross-patch fusion. The \emph{foundation-model} route, which has grown rapidly since 2024, is discussed in Section~\ref{sec:foundation}. Across routes, the progression from cubemap to ICOSAP (in bi-projection methods) and from ERP-plane tokens to spherical-domain tokens (in Transformer methods) points to a shared movement toward sphere-native representations.

\begin{table*}[tp]
\centering
\footnotesize
\caption{Representative panoramic depth estimation methods. Stage abbreviations follow the taxonomy of Section~\ref{sec:ch3}: PB = projection-based, DA = distortion-aware, SN = sphere-native, GT = geometry-aware tokenization, FMA = foundation-model adaptation.}
\label{tab:pano_depth}
\begin{tabular}{llccl}
\toprule
\textbf{Method} & \textbf{Venue} & \textbf{Year} & \textbf{Stage} & \textbf{Core Strategy} \\
\midrule
BiFuse \cite{Wang2020BiFuse} & CVPR & 2020 & PB & ERP$+$cubemap bi-fusion \\
UniFuse \cite{Jiang2021UniFuse} & RA-L & 2021 & PB & ERP$\leftarrow$cubemap uni-fusion \\
HRDFuse \cite{Ai2023HRDFuse} & CVPR & 2023 & PB & ERP$+$TP holistic-regional \\
Elite360D \cite{Ai2024Elite360D} & CVPR & 2024 & PB & ERP$+$ICOSAP, B2F attention \\
PanoFormer \cite{Shen2022PanoFormer} & ECCV & 2022 & DA & Spherical tangent tokens \\
SGFormer \cite{Zhang2025SGFormer} & TCSVT & 2025 & GT & Spherical prior decoder \\
Depth Anywhere \cite{Wang2024DepthAnywhere} & NeurIPS & 2024 & FMA & DA distillation $+$ random rot. \\
PanDA \cite{Cao2025PanDA} & CVPR & 2025 & DA$+$FMA & DAv2 LoRA $+$ M\"obius aug.\ $+$ EPNL \\
DA360 \cite{Jiang2025DA360} & arXiv & 2025 & FMA & DAv2 $+$ scale-inv. shift $+$ circ. pad \\
DA$^2$ \cite{Li2025DA2} & arXiv & 2025 & GT$+$FMA & SphereViT, 607K engine \\
DAP \cite{Lin2025DAP} & arXiv & 2025 & DA$+$FMA & DINOv3-L, 2M panoramas \\
\bottomrule
\end{tabular}
\end{table*}

\paragraph{Room layout and surface normal estimation.}
Panoramic layout estimation has moved through progressively more compact representations: from LayoutNet's~\cite{Zou2018LayoutNet} dense 2D boundary maps, through HorizonNet's~\cite{Sun2019HorizonNet} 1D representation that brought post-processing from dozens of seconds down to under 20\,ms, to LED$^2$-Net's~\cite{Wang2021LED2Net} horizon-depth formulation and LGT-Net's~\cite{Jiang2022LGTNet} geometry-aware Transformer. DOPNet~\cite{Shen2023DOPNet} further disentangles orthogonal planes with cross-scale distortion awareness, while AtlantaNet~\cite{Pintore2020AtlantaNet} relaxes the dominant Manhattan-world assumption to support non-orthogonal and curved walls. Layout estimation enjoys a structural advantage that other panoramic tasks do not, complete room visibility from a single image, but it is among the panoramic tasks that foundation models have so far left largely untouched, together with surface normal estimation. Surface normal estimation has received comparatively little attention as a standalone panoramic task: PanoNormal~\cite{huang2024PanoNormal} is among the few methods dedicated to panoramic normals, and normals otherwise appear as secondary outputs within multi-task frameworks such as HUSH~\cite{Lee2025HUSH} and Elite360M~\cite{Ai2024Elite360M}.

\subsection{Unified Scene Understanding}
\label{sec:unified}

The substantial overlap in distortion-handling requirements across depth, layout, normal, and segmentation motivates a single representation serving multiple panoramic tasks at once. The literature shows three generations of unified panoramic understanding, distinguished by the mechanism through which tasks are coupled. The first generation, represented by HoHoNet~\cite{Sun2021HoHoNet}, compressed the feature pyramid into a compact Latent Horizontal Feature (LHFeat) that exploits the vertical regularity of gravity-aligned indoor scenes, supporting depth, layout, and semantic estimation through separate lightweight decoders at 52\,FPS, but with no mechanism for cross-task information exchange. The second generation introduced explicit inter-task coupling: PanelNet~\cite{Yu2023PanelNet} represents panoramas as consecutive vertical panels with geometry embeddings and a Local2Global Transformer, while Elite360M~\cite{Ai2024Elite360M} adds a Cross-task Collaboration (CoCo) module that first extracts task-specific geometric and semantic features from a shared bi-projection representation and then integrates spatial context \emph{across} tasks, allowing depth discontinuities to inform semantic boundaries and vice versa. The third generation, marked by HUSH~\cite{Lee2025HUSH} at CVPR 2025, achieves unification through a physics-informed representational basis rather than architectural engineering: spherical harmonics (SH) basis functions, which form a natural orthogonal basis on the unit sphere, serve as queries in a hierarchical cross-attention module, and an SH basis index module selects, pixel-wise, which SH bases are most task-relevant for each of depth and surface normal. HUSH achieves state-of-the-art depth estimation on Stanford2D3D, Matterport3D, and Structured3D, while supporting surface normal and layout estimation from the same backbone, showing that the mathematical structure of spherical signals can substitute for task-specific architectural design. The three-generation arc, from data-driven feature compression through engineered cross-task coupling to physics-informed representational alignment, tracks the broader field-level drift from distortion-aware engineering toward sphere-native modeling.

\subsection{Foundation-Model-Assisted Understanding}
\label{sec:foundation}

Foundation models pretrained on perspective imagery at massive scale have reshaped panoramic dense prediction since 2024, but their adaptation to $360^{\circ}$ input has to bridge gaps in field of view, geometric distortion, and, for SAM-family models, the semantic understanding that class-agnostic pretraining does not provide.

\paragraph{SAM adaptation for panoramic segmentation.}
GoodSAM~\cite{Zhang2024GoodSAM} and GoodSAM++~\cite{Zhang2024GoodSAMplus} used SAM indirectly as a pseudo-label teacher, generating ensemble logits from perspective crops to supervise a separate panoramic student via knowledge distillation. OmniSAM~\cite{Zhong2025OmniSAM} integrates SAM2 directly: it reframes the panorama as a sequence of overlapping patches (analogous to video frames) and exploits SAM2's memory mechanism for cross-patch spatial correspondence, fine-tunes the SAM2 image encoder with LoRA, and uses a FoV-based prototypical adaptation module with dynamic pseudo-label updating for target-domain alignment. It reaches 79.06\% mIoU on SPin8-to-SPan8 indoor Pin2Pan UDA (a $+10.22$ percentage-point gain over prior UDA baselines) and 62.46\% mIoU on CS13-to-DP13 outdoor UDA ($+6.58$ pp). PanoSAMic~\cite{Chamseddine2026PanoSAMic} freezes the SAM encoder and adds multi-modal fusion (RGB, depth, normals) with a dual-view pipeline that processes the original panorama and a horizontally shifted copy in parallel before merging their feature maps, mitigating ERP boundary discontinuity and reaching state-of-the-art mIoU on Stanford2D3DS and Matterport3D across multiple input modalities. A common pattern is that SAM adaptation for panoramas is not a simple fine-tuning exercise: each method introduces substantial engineering (memory-based stitching, dual-view fusion, or multi-modal attention) to absorb the mismatch between perspective pretraining and panoramic deployment.

\paragraph{Foundation models for panoramic depth.}
The shared problem here is narrow: a depth foundation model pretrained on perspective images already predicts excellent relative depth, but its convolution and attention assume a planar grid, so it mishandles ERP distortion and the longitudinal seam. Two branches address this with opposite cost profiles (Table~\ref{tab:pano_depth}, bottom rows). The \emph{adaptation branch} keeps Depth Anything V2 (DAv2)~\cite{Yang2024DepthAnythingV2} essentially intact and adds only the minimum needed for $360^{\circ}$, a distortion-robust training signal and seam-aware continuity; its three representative methods differ mainly in which knobs they turn, distillation from a perspective teacher with cube-projection pseudo-labels and random rotation (Depth Anywhere~\cite{Wang2024DepthAnywhere}), LoRA fine-tuning with M\"obius augmentation and an equator-aware normalization loss (PanDA~\cite{Cao2025PanDA}), or just a learned global shift plus circular padding (DA360~\cite{Jiang2025DA360}), and together they show that surprisingly small changes recover much of the lost accuracy. The \emph{construction branch} instead builds panorama-specific architecture and corpora around the pretrained weights, its members varying mainly in how spherical geometry is injected: DA$^2$~\cite{Li2025DA2} lets image features cross-attend to spherical-coordinate embeddings (SphereViT, $\sim$607K panoramas, average 38\% AbsRel gain over the strongest zero-shot baseline), DAP~\cite{Lin2025DAP} scales the same idea to DINOv3-Large on 2M panoramas, and UniK3D~\cite{Piccinelli2025UniK3D} and Depth Any Camera~\cite{Guo2025DAC} fold the camera model itself into the network, the former as a learned superposition of spherical harmonics, the latter as a unified representation spanning pinhole, fisheye, and panoramic inputs. The contrast is the lesson: adaptation is cheap but inherits perspective inductive biases, whereas construction yields better geometry at the price of orders-of-magnitude more panoramic data. Either way the same gap persists: no panoramic depth foundation model has yet been pretrained from scratch on spherical data; every method above initializes from perspective-pretrained weights.

\subsection{Open-World Panoramic Understanding}
\label{sec:openworld}

Moving from closed-set to open-world understanding introduces challenges that go beyond those familiar from the perspective literature. Open-vocabulary recognition, out-of-distribution (OOD) detection, and zero-shot generalization are shared concerns, but the panoramic setting adds three further complications. First, the FoV gap between perspective-pretrained VLMs and $360^{\circ}$ test imagery is a geometric domain shift that alters object appearance as a function of latitude. Second, ERP distortion disproportionately affects novel categories that have never had the chance to learn distortion compensation from training exposure. Third, seam discontinuity can split a novel object across the ERP boundary into two unrelated predictions. In the other direction, the $360^{\circ}$ contextual completeness (simultaneous observation of all surrounding objects and their spatial relationships) represents an untapped advantage for open-world reasoning that no existing method exploits.

The open-world panoramic literature remains nascent and consists of a handful of task-defining works. OPS~\cite{Zheng2024OPS} at ECCV 2024 formalized Open Panoramic Segmentation, a setting in which models are trained under open-vocabulary supervision on perspective images and evaluated zero-shot on panoramas. The same paper introduced OOOPS, which combines a Deformable Adapter Network with Random Equirectangular Projection (RERP) augmentation to bridge the FoV and distortion gaps. PanOoS~\cite{Duan2025PanOoS} defined Panoramic Out-of-Distribution Segmentation and proposed POS, a CLIP-based method with Prompt-based Restoration Attention and Bilevel Prompt Distribution Learning that improves AuPRC by 34.25\% and reduces FPR95 by 21.42\% on the DenseOoS benchmark relative to pinhole-OoS baselines applied to panoramic input, confirming that panorama-specific adaptation is essential for reliable OOD detection. JOPP-3D~\cite{Inuganti2026JOPP3D} extends the paradigm to joint open-vocabulary segmentation across panoramic images and 3D point clouds via tangential decomposition and depth-based 3D-to-2D correspondence. With only a handful of papers in the area, these works serve as foundational benchmarks and baselines rather than mature solutions. Open-world panoramic work is young for a structural reason: its prerequisites (panoramic foundation models, panoramic VLM alignment, panoramic OOD benchmarks) have only become available in the last one to two years.

\subsection{Synthesis: Established Results and Open Gaps}
\label{sec:ch4_discussion}

Pulling the dense-prediction literature together, three claims now rest on strong multi-study support: distortion-aware architectures outperform direct perspective transfer, unsupervised domain adaptation mitigates panoramic annotation scarcity, and foundation-model distillation improves panoramic depth. Two more are only partially supported: that geometry-aware representations beyond plain ERP (SH bases in HUSH, ICOSAP sampling in Elite360D, SphereViT in DA$^{2}$) deliver genuinely better geometric accuracy rather than merely better sphere-aware sampling, and that unified multi-task learning matches single-task specialists. Three remain aspirational, plausible but not yet validated: that Mamba will supersede Transformers for PASS, that $360^{\circ}$ contextual completeness aids open-world recognition, and that sphere-native pretraining would outperform perspective-initialized adaptation. This last point names the field's central unfilled gap, that no panoramic foundation model has yet been pretrained from scratch on spherical data (Section~\ref{sec:foundation}); foundation models have moved fastest on depth while leaving layout and surface-normal estimation largely untouched, and in the open-world setting the FoV gap has absorbed most of the effort while seam-aware prediction and the deliberate use of contextual completeness remain unaddressed.

These observations crystallize into six concrete \emph{task-level} gaps, distinct from the five \emph{evaluation-protocol} gaps of Section~\ref{sec:eval_critique} and the six \emph{methodological} directions of Section~\ref{sec:open_problems}: (i) the near-absence of dedicated panoramic normal estimation; (ii) the lack of foundation-model-driven layout estimation; (iii) unaddressed seam-aware open-world prediction; (iv) unexploited $360^{\circ}$ contextual completeness for novel-class recognition; (v) the missing sphere-native foundation model pretrained without perspective initialization; and (vi) the small, fragmented state of densely annotated outdoor panoramic data relative to both mature perspective driving benchmarks and indoor panoramic datasets.

Having surveyed pixel-level panoramic understanding from closed-set through open-world, we next examine how vision-language models extend panoramic perception beyond dense prediction into multimodal reasoning and spatial understanding.


\section{Panoramic Vision-Language Understanding and Spatial Reasoning}
\label{sec:ch5}

This section turns to tasks where the output is natural language and the model must reason about what it sees. Panoramic images are an appealing substrate for vision-language models (VLMs): the entire surrounding scene arrives in a single tensor, with no blind spots and none of the cross-camera fusion cost or view-boundary inconsistency that multi-view rigs such as BEVFormer~\cite{Li2022BEVFormer} must manage~\cite{Dense360_2025}. The evidence that this helps language grounding is concrete: DeepPanoContext~\cite{Zhang2021DeepPanoContext} used the wider field of view as scene context for a relation-based graph model of 3D room layout, PLM~\cite{Fan2026PLM} reported on its PanoVQA benchmark that a single cylindrical panorama beats a six-camera multi-view input even with fewer total pixels, and PanoGrounder~\cite{PanoGrounder2025} reached state-of-the-art 3D grounding on ScanRefer and Nr3D by pairing panoramic renderings with pretrained 2D VLMs; panoramas are likewise the standard input for vision-and-language navigation, going back to Room-to-Room and the Matterport3D simulator~\cite{Anderson2018R2R}. Capturing the whole scene, however, is not the same as understanding it: the literature progresses through increasingly deep accommodation, from treating ERP as an ordinary photograph, through geometric priors injected at the encoding and reasoning stages, to a still-unrealized sphere-native VLM (Section~\ref{sec:ch5-spatial}).

\subsection{Panoramic Vision-Language Tasks: VQA, Captioning, and Grounding}
\label{sec:ch5-dense}

Panoramic VQA differs from the dense prediction tasks of Section~\ref{sec:scene_understanding} in that the output is free-form text and the model must jointly locate and describe entities inside a distorted representation. VQA $360^{\circ}$~\cite{Chou2020VQA360} introduced the first benchmark, about 17K question-answer pairs over Stanford2D-3D and Matterport3D, and found that projecting into six cubemap faces with multi-level attention beat feeding ERP to a standard VQA model directly, an early signal that distortion handling is a prerequisite. Pano-AVQA~\cite{Yun2021PanoAVQA} extended the setting to audio-visual QA on $360^{\circ}$ video with spherical spatial embeddings that encode relative rather than principal-axis orientation, since a panorama has no canonical front. Captioning raises a parallel difficulty: a $360^{\circ}$ image holds far more content than a perspective photograph, so a single caption tends to omit entities or collapse into generality. Maeda et al.~\cite{Maeda2023QuIC} condition the caption on a text query (QuIC), and the 360DVD~\cite{Wang2024_360DVD} pipeline captions per perspective view and fuses with an LLM rather than captioning ERP directly.

Dense360~\cite{Dense360_2025} is the largest effort in this direction, contributing 160K panoramas with 5M entity-level captions, 1M referring expressions, and 100K entity-grounded scene descriptions, each anchored to a localized entity rather than to one global caption and carrying a reliability score. On its Dense360-Bench, open-source MLLMs lag well behind their perspective-input performance, which the authors attribute to the near-absence of ERP data in pretraining corpora; the accompanying Dense360VLM (Qwen2.5-VL-3B with the ERP-RoPE encoding of Section~\ref{sec:ch5-spatial}) reaches 51.78 captioning recall and 76.81 grounding mask IoU against the SA2VA-4B baseline (47.80 and 74.39), with ERP-RoPE alone contributing +5.92 and +16.38. Two further works push beyond captioning: PanoAffordanceNet~\cite{zhu2026PanoAffordance} moves grounding from objects to functional regions through a frequency-domain distortion modulator and a densification head that recovers continuous affordance regions from sparse keypoints, releasing a dedicated panoramic affordance dataset, while 360-R1~\cite{Zhang2025OmniVQA360R1} applies GRPO with structured rewards for reasoning consistency, correctness, and format, reporting a roughly 6\% gain over a Qwen2.5-VL baseline on spatial reasoning.

\subsection{Geometry-Aware Encoding and Spatial Reasoning for Panoramic VLMs}
\label{sec:ch5-spatial}

The tokenization problem this subsection starts from is the same ERP geometry described in Section~\ref{sec:geometry}; what is new is not the geometry but its consequences in the vision-language setting, for two reasons. VLMs are pretrained almost entirely on perspective image-text pairs, so they have essentially never seen ERP at scale, and their target output, language-grounded spatial reasoning, is far less tolerant of geometric error than a per-pixel label map. The encoding problems below are therefore inherited from Sections~\ref{sec:geometry} and~\ref{sec:scene_understanding}, whereas the reasoning failures later in this subsection are genuinely specific to VLMs. Concretely, standard ViT tokenization rests on two assumptions that do not hold on ERP.
Patches are treated as having uniform area, whereas polar ERP patches subtend a much smaller solid angle than equatorial ones.
Positional encoding is treated as planar, whereas the left and right edges of an ERP image are adjacent on the sphere but receive maximally distant position indices.
360Bench~\cite{Tran2026Free360} quantifies the resulting gap.
The strongest MLLM evaluated (Gemini Pro 2.5) reaches 46.5\% overall accuracy against 86.3\% for humans, and the projection format strongly affects accuracy: cubemap input outperforms ERP by up to $+14.1$\% on projection-distorted subtasks (LLaVA-CoT on PP-IR), while ERP outperforms cubemap by up to $+14.5$\% on spatial-reasoning subtasks (Gemini 2.5 Flash on SR-OV)~\cite{Tran2026Free360}.

ERP-RoPE~\cite{Dense360_2025} (Section~\ref{sec:geo_token}) is realized at the VLM level by modifying only the width-axis component of the Qwen2.5-VL M-RoPE encoding (circular in longitude, latitude-scaled); the Dense360 ablation shows that this single modification closes most of the captioning gap between the front and back of the panorama, directly addressing the seam discontinuity.
PLM~\cite{Fan2026PLM} takes a different route at the attention level and supplies a plug-and-play Panorama Sparse Attention module that adapts attention patterns to the $360^{\circ}$ structure and can be inserted into an existing pinhole-based VLM without full retraining.
Two further strategies are training-free.
Omni-CoT~\cite{Yang2025ODIBench} adds a three-step reasoning chain (viewpoint-guided answering, crop-cue grounding, and response refinement), with the original paper reporting consistent gains over direct answering and zero-shot CoT, especially on spatial-level subtasks.
Free360~\cite{Tran2026Free360} builds scene graphs using entity-centered spherical rotations, so that each entity of interest is re-centered before analysis, compensating for ERP distortion without touching the model parameters, reporting up to 22.9\% gain on individual subtasks and 7.3\% overall on 360Bench.

\begin{table}[!htbp]
\centering
\caption{Panoramic spatial reasoning and understanding benchmarks. \emph{3D}: 3D ground-truth annotations available; \emph{NS}: negative sampling for hallucination testing.}
\label{tab:ch5-bench}
\footnotesize
\setlength{\tabcolsep}{2pt}
\begin{tabular}{@{}lcrrccc@{}}
\toprule
\textbf{Benchmark} & \textbf{Yr} & \textbf{\#Img} & \textbf{\#QA} & \textbf{Scene} & \textbf{3D} & \textbf{NS} \\
\midrule
VQA 360\textdegree{}~\cite{Chou2020VQA360} & '20 & 1,490 & 17K  & In  & -- & -- \\
Pano-AVQA~\cite{Yun2021PanoAVQA} & '21 & 5.4K$^*$  & 52K  & Mix & -- & -- \\
OSR-Bench~\cite{Dongfang2025OSRBench}       & '25 & 4.1K  & 153K & In  & -- & \cmark \\
Dense360-B~\cite{Dense360_2025}             & '25 & 1,279 & 6K   & Mix & -- & -- \\
ODI-Bench~\cite{Yang2025ODIBench}           & '25 & 2K    & 4.3K & Mix & -- & -- \\
PanoEnv~\cite{Pan2026PanoEnv}                  & '26 & --    & 14.8K & Mix & \cmark & -- \\
360Bench~\cite{Tran2026Free360}             & '26 & 643   & 6.1K & Mix & -- & -- \\
\bottomrule
\end{tabular}
\par\smallskip
{\raggedright\scriptsize $^*$Video clips, not still images.\par}
\end{table}

Table~\ref{tab:ch5-bench} summarizes the rapidly growing benchmark landscape.
OSR-Bench~\cite{Dongfang2025OSRBench} contributes 153K QA pairs with negative sampling designed to test hallucination robustness, and both GPT-4o and Gemini 1.5 Pro perform poorly on it under zero-shot evaluation.
PanoEnv~\cite{Pan2026PanoEnv} moves closer to genuine 3D physical reasoning by targeting metric volumes, distances, and viewpoint meta-reasoning from a single monocular ERP image.
It uses geometry-grounded ground truth from a simulation engine as the GRPO reward signal, which avoids a common failure mode of reinforcement learning with LLM-generated answers, namely rewarding hallucinated reasoning traces.
The training curriculum also matters: simultaneous training on structured and open-ended questions induces catastrophic forgetting on the structured tasks, whereas a two-stage schedule (structured first, then open-ended) reaches 52.93\% total accuracy with a 7B backbone, ahead of 32B baselines, with open-ended accuracy rising 132\% in relative terms (from 6.39\% to 14.83\%).
ODI-Bench~\cite{Yang2025ODIBench} highlights a complementary failure: on non-egocentric tasks such as allocentric orientation and scene simulation, model accuracy sits only marginally above a blind baseline, which supports the claim at the start of this section that current MLLMs do not extract the immersive spatial information unique to omnidirectional images.

Three specific causes of this gap can be identified from the broader VLM literature.
Current panoramic VLMs do not receive metric depth supervision; SpatialVLM~\cite{Chen2024SpatialVLM} showed that quantitative spatial reasoning in the perspective domain requires depth-annotated training data, which panoramic VLMs currently lack.
They also do not use a dedicated geometric encoder; Spatial-MLLM~\cite{Wu2026SpatialMLLM} reaches state-of-the-art spatial reasoning in the perspective domain by separating semantic and geometric features into a dual-encoder design, and no panoramic VLM has adopted this separation.
Finally, they are camera-agnostic; recent work on camera-aware MLLMs~\cite{CameraAwareMLLM2026} demonstrates that simply resizing the input image shifts 3D localization in systematic ways, and ERP's latitude-dependent scaling is itself a spatially varying distortion that calls for explicit camera modeling rather than implicit learning.

\subsection{Toward Panoramic Multimodal Foundation Models}
\label{sec:ch5-future}

No current panoramic vision-language method is truly panorama-native. Existing work occupies the first three rungs of the ladder previewed at the start of this section, zero adaptation (a standard MLLM treats ERP as an ordinary photograph), encoding-level adaptation (ERP-RoPE or Panorama Sparse Attention inject geometric priors into an otherwise planar tokenizer), and reasoning-level adaptation (Omni-CoT and Free360 compensate at inference through perspective decomposition or entity-centered rotation), while a fourth rung, a visual encoder operating directly on spherical geometry within a unified vision-language framework, remains unexplored. Reaching it would require pieces not yet in place: equal-area tokenization on the viewing sphere rather than equal-rectangle tokens on ERP; a dual-encoder design that separates semantic and geometric features in the spirit of Spatial-MLLM~\cite{Wu2026SpatialMLLM} but with spherical priors; pretraining data well beyond the largest current corpus (Dense360~\cite{Dense360_2025}, 160K panoramas, orders of magnitude below CLIP-scale image-text data); and projection-aware modeling that treats the ERP mapping as a known prior rather than assuming it will be recovered implicitly. Progress has been rapid, from distortion-compensating VQA~\cite{Chou2020VQA360} through geometry-aware encoding~\cite{Dense360_2025} and reinforcement-learning-enhanced reasoning~\cite{Pan2026PanoEnv} to explicit panorama-language modeling~\cite{Fan2026PLM}, but connecting these threads to the sphere-native representations of Section~\ref{sec:ch3} within one model that perceives, encodes, and reasons in spherical geometry end to end remains open.


\section{Dynamic Panoramic Perception: Video, Tracking, and Embodied Scenes}
\label{sec:ch6}

Sections~\ref{sec:scene_understanding} and~\ref{sec:ch5} treated panoramic scene understanding as a single-frame problem. Real-world panoramic sensing is inherently temporal, and moving from static to dynamic perception is not a matter of running video methods on ERP frames as if nothing had changed. This section examines the spherical-geometry challenges that arise in the temporal domain and surveys the small body of work that addresses them. The dedicated literature is thin: at the time of writing, fewer than ten tracking or segmentation methods, a handful of optical-flow approaches, and a small set of video-level benchmarks make up the entire landscape, an asymmetry we surface as one of the field's central open problems (Section~\ref{sec:open_problems}).

\begin{figure*}[t!]
\centering
\includegraphics[width=0.85\linewidth]{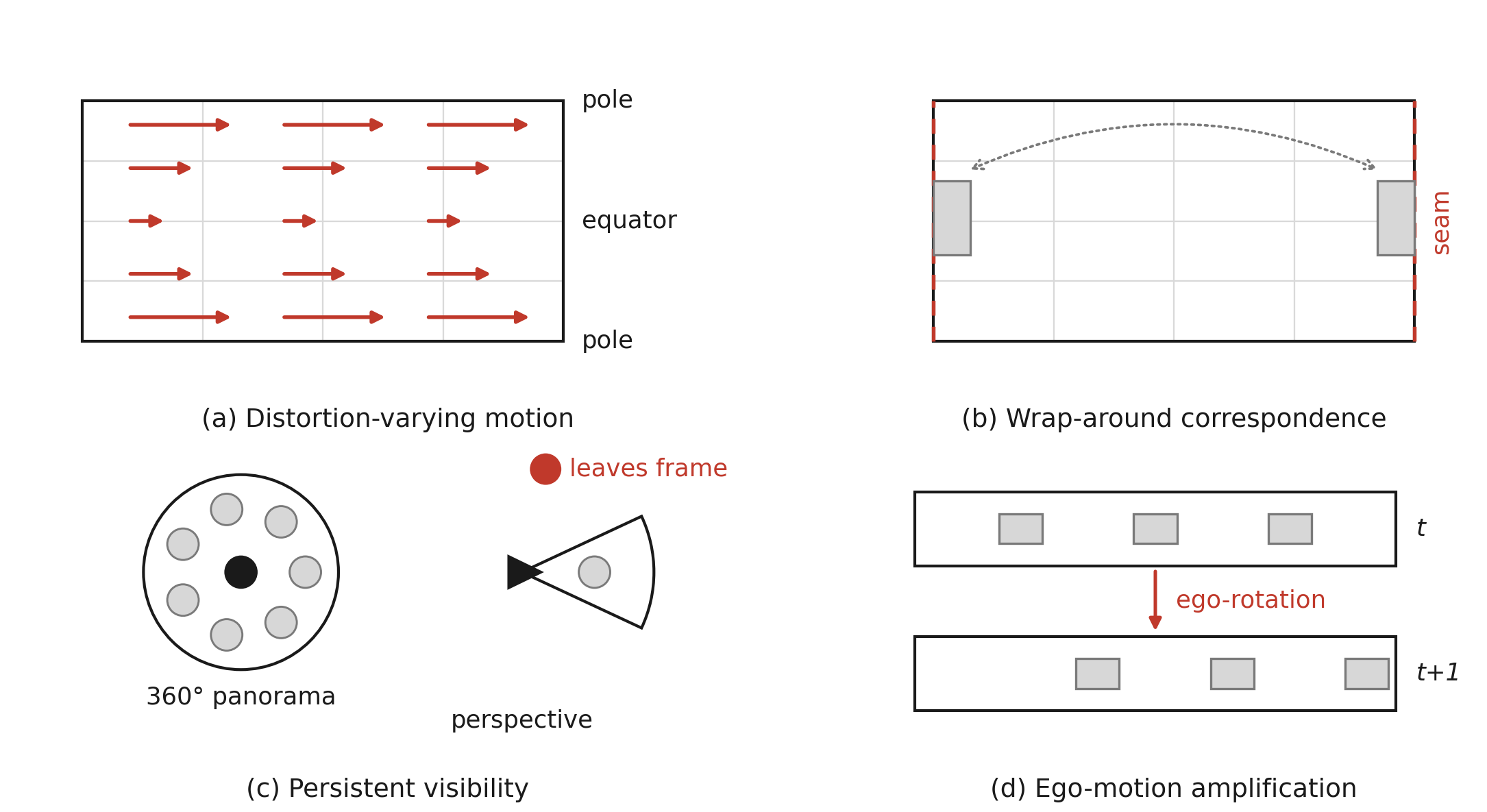}
\caption{Four structural challenges that distinguish dynamic panoramic perception from perspective video. (a)~Identical 3D motion induces pixel-level optical flow whose magnitude grows toward the poles, violating the translation invariance assumed by standard flow estimators. (b)~An object crossing the equirectangular seam is split into two fragments at opposite image borders. (c)~In a $360^{\circ}$ panorama, surrounding objects remain in view (occlusion aside), so identity must be maintained continuously rather than through entry/exit events. (d)~On a mobile platform, even a small ego-rotation shifts the entire scene globally.}
\label{fig:dynamic_challenges}
\end{figure*}

\subsection{Challenges of Temporal Panoramic Perception}
\label{sec:ch6_challenges}

Four structural properties separate dynamic panoramic perception from perspective video analysis (Fig.~\ref{fig:dynamic_challenges}). The first is distortion-varying motion: the same 3D translation produces pixel-level flow whose magnitude varies with latitude on ERP, which violates the translational invariance that standard flow estimators assume~\cite{Shi2023PanoFlow}. The second is wrap-around correspondence: when a tracked object crosses the ERP seam, its bounding box splits into two fragments at opposite image borders, and region-of-interest features built on a single connected box become unreliable. Both 360VOT~\cite{Huang2023360VOT} and PanoVOS~\cite{Yan2024PanoVOS} report this as one of the dominant failure modes of off-the-shelf trackers and segmenters. The third is persistent visibility: in perspective video, objects leave the frame and return, but in panoramic video surrounding objects stay within the image (occlusion aside), so identity has to be maintained continuously rather than through entry/exit events~\cite{Luo2025OmniTrack}. The fourth is ego-motion amplification: on a mobile platform, even a small sensor rotation shifts the entire ERP scene globally, an effect that narrow-FoV cameras largely avoid. OmniTrack++~\cite{Luo2025OmniTrackPlus} reports that strong perspective MOT baselines degrade sharply when transferred to robot-centric panoramic settings; the cross-dataset comparison between DanceTrack and JRDB they cite mixes domain, scene, and task setting and should be read as an order-of-magnitude motivator rather than a controlled measurement. These are qualitative differences from perspective-video assumptions, not just quantitative ones.

\subsection{Tracking, Segmentation, and Motion Estimation}
\label{sec:ch6_methods}

The methods in this subsection are best read as targeted responses to the four challenges of Section~\ref{sec:ch6_challenges}. Distortion-aware target representations (BFoV, gnomonic re-projection) and seam-aware propagation answer distortion-varying motion and wrap-around correspondence; trajectory- and memory-based association answer persistent visibility and ego-motion amplification. A recurring limitation, developed in Section~\ref{sec:ch6_gaps}, is that almost all of them address only the \emph{spatial} side of these challenges and inherit their \emph{temporal} components unchanged from perspective video.

\paragraph{Single-object tracking.}
360VOT~\cite{Huang2023360VOT} is the first benchmark for omnidirectional SOT. It introduces the bounding field-of-view (BFoV) target representation together with sphere-aware evaluation metrics (dual success, dual precision, angle precision), provides 120 test sequences with up to 113K frames, and reports that 20 state-of-the-art perspective trackers degrade substantially on distortion and seam-crossing scenarios. Its extension 360VOTS~\cite{Xu2025360VOTS} unifies tracking and segmentation using an extended BFoV (eBFoV) for search regions and adds 360VOS, a 290-sequence set with dense mask annotations that can be used to lift any off-the-shelf tracker to $360^{\circ}$ inputs via gnomonic projection. Peng et al. ~\cite{peng2025Robust360} propose a dynamic gnomonic projection coupled with trajectory prediction for recovery after tracking drift. All three works stay within the projection-based paradigm and do not build sphere-native temporal representations.

\paragraph{Multi-object tracking.}
OmniTrack~\cite{Luo2025OmniTrack} is, to our knowledge, the first unified MOT framework designed for $360^{\circ}$ imagery. It combines a CircularStatE module that normalizes features across the panoramic FoV, FlexiTrack Instances that inject trajectory cues into localization and association, and a Tracklet Management module that switches between end-to-end and tracking-by-detection paradigms. The original OmniTrack reaches 26.92\% HOTA on JRDB~\cite{Martin2021JRDB} (+3.43 over baseline) and 23.45\% on the newly introduced QuadTrack dataset (+6.81 over baseline). Its extension OmniTrack++~\cite{Luo2025OmniTrackPlus} replaces CircularStatE with a DynamicSSM block and adds an ExpertTrack Memory that consolidates appearance cues through a mixture-of-experts design; relative to the original OmniTrack, OmniTrack++ reports HOTA gains of $+3.94$ on JRDB and $+15.03$ on QuadTrack; on QuadTrack, its end-to-end variant reaches $34.90$ HOTA and its tracking-by-detection variant reaches $36.08$ HOTA, both above the original OmniTrack. JRDB-PanoTrack~\cite{Le2024JRDBPanoTrack} complements this line of work by providing open-world panoptic segmentation and tracking annotations on $360^{\circ}$ robotic imagery, with up to 245 masks per panoramic frame in the densest scenes.

\paragraph{Video object segmentation.}
PanoVOS~\cite{Yan2024PanoVOS} is the first long-term, instance-level panoramic VOS benchmark, with 150 videos and 19K annotated instance masks. The paper evaluates 15 off-the-shelf VOS models and finds that all of them fail to handle the pixel-level content discontinuity introduced by the ERP seam. The companion method, PSCFormer, exploits the geometric fact that the left and right boundaries of an ERP image are contiguous on the sphere, which enables coherent mask propagation for objects crossing the seam. On the data side, Leader360V~\cite{Zhang2026Leader360V} provides the first large-scale $360^{\circ}$ video dataset (10K+ videos) for instance segmentation and tracking, with an automatic annotation pipeline that combines pre-trained 2D segmentors, SAM2, and LLM-based verification, which directly addresses the labeled-data bottleneck.

\paragraph{Optical flow and temporal understanding.}
PanoFlow~\cite{Shi2023PanoFlow} proposes cyclic flow estimation, which exploits the boundaryless topology of the sphere to convert large seam-crossing displacements into two smaller complementary ones, and reports a 55.5\% EPE reduction over the best prior result on OmniFlowNet. Panoramic video saliency has progressed from spherical-crown CNNs~\cite{Zhang2018Saliency360} to panoramic ViTs~\cite{Yun2022PAVER} and audiovisual models that exploit ambisonic audio~\cite{Zhang2023PAVSOD,Li2023PanoVideoSOD}. Panoramic activity recognition~\cite{Han2022PanoramicPAR,Cao2024AdaFPP} targets multi-granularity behavior understanding (individual actions, social group activities, and global activity) over the full $360^{\circ}$ scene, and Pano-AVQA~\cite{Yun2021PanoAVQA} remains the only benchmark for grounded question answering on panoramic video.

\paragraph{Foundation models for panoramic video.}
The intersection of panoramic video with video foundation models has only just begun to open up. OmniSAM~\cite{Zhong2025OmniSAM} and SAP~\cite{Jiang2026SAP} (introduced in Sections~\ref{sec:geo_token} and~\ref{sec:foundation}) repurpose SAM2's memory mechanism for spatial consistency on panoramic \emph{images}, decomposing the panorama into patch or perspective-view sequences and feeding them to SAM2 as if they were video frames; they do not process genuine video. The first systematic adaptation to true panoramic \emph{video sequences} is the very recent PanoSAM2~\cite{Xiao2026PanoSAM2}, which appeared only in 2026 and so far is confined to single-object 360VOS with seam-aware decoding and long-short memory adjustments. Multi-object tracking, identity persistence under ego-motion, and on-sphere motion estimation remain untouched by foundation-model approaches.

\subsection{Summary and Open Gaps}
\label{sec:ch6_gaps}

The dominant pattern across this small body of work is \emph{spatial} sphere-awareness combined with \emph{temporal} planarity: methods address the spatial side of ERP (distortion, seam handling, BFoV representations) but inherit their temporal components (Kalman filters, memory banks, recurrent modules) unchanged from perspective video. To our knowledge, no published method formulates temporal association or motion prediction natively on the sphere, for instance through geodesic-distance state estimation or spherical-harmonic motion. The same thinness marks the foundation-model frontier, where adaptation to genuine panoramic video has only begun with PanoSAM2~\cite{Xiao2026PanoSAM2} and remains limited to single-object segmentation, and the embodied setting, where platforms such as Habitat~\cite{Savva2019Habitat} and the JRDB ecosystem~\cite{Martin2021JRDB,Le2024JRDBPanoTrack} supply panoramic observations that most agents still process without exploiting spherical structure in time~\cite{Zheng2025PANORAMA}.

This matches the survey's overall picture: the shift from distortion-aware engineering toward sphere-native modeling is least advanced in the temporal domain, where dynamic perception is still at the first of the paradigm transitions that static understanding has already passed through. Closing it is among the clearer priorities for the next generation of panoramic scene understanding.

\section{Datasets, Benchmarks, and Evaluation Protocols}
\label{sec:datasets}

This section surveys the data infrastructure behind the evolution traced in Sections~\ref{sec:ch3}--\ref{sec:ch6}: the benchmarks on which methods are trained, the datasets against which they are measured, and the evaluation protocols that decide what counts as progress.

We organize the landscape into four benchmark families (Table~\ref{tab:benchmarks}), and then examine five dimensions along which current protocols fall short of testing genuine spherical understanding.

\subsection{Benchmark Landscape}
\label{sec:bench_landscape}

\paragraph{Static scene understanding.}
The core benchmarks for panoramic segmentation, depth estimation, and layout prediction remain predominantly indoor.
Stanford2D3D~\cite{Armeni2017Stanford2D3D} provides 1,413 equirectangular projection (ERP) panoramas across 6 areas with 13-class pixel-level annotations, and has become the de facto standard for indoor panoramic segmentation.
Matterport3D~\cite{Chang2017Matterport3D} offers 10,800 panoramic views from 90 building-scale scenes with 2D and 3D semantic labels, and also serves as the visual backbone for vision-and-language navigation (VLN) research.
Structured3D~\cite{Zheng2020Structured3D}, the largest densely multi-task-annotated synthetic panoramic dataset, renders 21,835 rooms from 3,500 professionally designed houses with dense annotations for segmentation, depth, normal, and layout tasks.
3D60~\cite{Zioulis2018OmniDepth} aggregates stereo panoramic renders from Matterport3D, Stanford2D3D, and SunCG for depth and surface normal estimation.
For outdoor driving scenarios, DensePASS~\cite{Ma2021DensePASS} releases 100 densely annotated panoramas with 19 Cityscapes-compatible classes for evaluation, while WildPASS~\cite{Yang2021WildPASS} releases a 500-panorama densely annotated evaluation set covering 25 cities (8 navigation-relevant classes); the larger WildPASS2K extension adds 2,000 unlabeled panoramas from 40 cities for unsupervised adaptation.
SynPASS~\cite{Zhang2024Trans4PASS} adds 9,080 synthetic outdoor panoramas across cloudy, foggy, rainy, sunny, and day/night conditions for synthetic-to-real adaptation.
SUN360~\cite{Xiao2012SUN360}, an early milestone with tens of thousands of panoramas spanning 80 scene categories, provides scene-level category labels but lacks pixel-wise semantic annotations.
Overall, this family is characterized by small annotation scale (Stanford2D3D has roughly 1,400 labeled panoramas versus ADE20K's 25K perspective images), heavy indoor bias, and exclusive reliance on ERP format.

\paragraph{Open-world and safety.}
The Open Panoramic Segmentation (OPS) task~\cite{Zheng2024OPS} formalizes zero-shot, open-vocabulary panoramic segmentation by training on pinhole images and evaluating on panoramic targets, using existing benchmarks (WildPASS, Stanford2D3D, Matterport3D) rather than introducing new annotations.
PanOoS~\cite{Duan2025PanOoS} extends panoramic perception to safety-critical out-of-distribution detection with two dedicated benchmarks: DenseOoS, which curates outlier placements in complex panoramic backgrounds, and QuadOoS, captured by a quadruped robot with a panoramic annular lens (PAL).

\paragraph{Multimodal and reasoning.}
A recent wave of benchmarks targets vision-language understanding in panoramic settings.
Pano-AVQA~\cite{Yun2021PanoAVQA} pioneered grounded audio-visual question answering on 5.4K panoramic video clips with spherical spatial and audio-visual relation QAs.
Since 2025, several panoramic VQA benchmarks have appeared: OSR-Bench~\cite{Dongfang2025OSRBench} evaluates spatial reasoning with 153K QA pairs grounded in cognitive maps; OmniVQA/360-R1~\cite{Zhang2025OmniVQA360R1} tests object identification, attribute analysis, and spatial reasoning in panoramic imagery; Dense360~\cite{Dense360_2025} introduces the largest panoramic vision-language dataset (160K panoramas, 5M entity captions) for captioning and grounding, with the accompanying Dense360-Bench split used for evaluation; and PanoEnv~\cite{Pan2026PanoEnv} builds geometry-grounded VQA from simulation data for 3D spatial intelligence assessment via reinforcement learning.
Each benchmark defines its own task taxonomy, QA format, and evaluation metrics, reflecting an active but fragmented evaluation landscape.

\paragraph{Video and tracking.}
360VOT~\cite{Huang2023360VOT} established the first omnidirectional single-object tracking benchmark, with 120 sequences and 113K frames, and introduced the bounding field-of-view (BFoV) representation together with sphere-tailored metrics (dual success, dual precision, and angle precision).
360VOTS~\cite{Xu2025360VOTS} extends this to video object segmentation, contributing the 360VOS dataset of 290 sequences with dense pixel-level masks in 62 object categories.
PanoVOS~\cite{Yan2024PanoVOS} provides 150 high-resolution panoramic videos with 19K instance masks, explicitly targeting content discontinuity at ERP boundaries.
For multi-object tracking, the primary evaluation beds are QuadTrack, the panoramic MOT benchmark captured by a quadruped robot and released with OmniTrack~\cite{Luo2025OmniTrack}, and EmboTrack, the umbrella benchmark released with OmniTrack++~\cite{Luo2025OmniTrackPlus} that subsumes QuadTrack together with BipTrack, the latter captured by a bipedal wheel-legged robot.
Leader360V~\cite{Zhang2026Leader360V} (described in detail in Section~\ref{sec:ch6_methods}) marks a qualitative leap in data scale, made possible by a foundation-model-assisted automatic annotation pipeline.

\begin{table*}[tp]
\centering
\caption{Major panoramic benchmarks used in Sections~\ref{sec:ch3}--\ref{sec:ch6}. Scale reports labeled samples unless noted. R = Real, S = Synthetic, M = Mixed. Status: \cmark{} = data and code publicly released as of submission; $\ddagger$ = announced as forthcoming in the source paper (verify current release status).}
\label{tab:benchmarks}
\footnotesize
\setlength{\tabcolsep}{3pt}
\resizebox{\textwidth}{!}{%
\begin{tabular}{@{}llllrlcl@{}}
\toprule
\textbf{Name} & \textbf{Year} & \textbf{Domain} & \textbf{Task(s)} & \textbf{Scale} & \textbf{R/S} & \textbf{Format} & \textbf{Status} \\
\midrule
\multicolumn{8}{l}{\textit{Static Scene Understanding}} \\
SUN360~\cite{Xiao2012SUN360}            & 2012 & In/Outdoor  & Scene recog.             & $\sim$67K panos         & R & ERP     & \checkmark \\
Stanford2D3D~\cite{Armeni2017Stanford2D3D} & 2017 & Indoor   & Seg., depth, normal      & 1,413 panos             & R & ERP     & \checkmark \\
Matterport3D~\cite{Chang2017Matterport3D}  & 2017 & Indoor   & Seg., depth, VLN         & 10,800 pano views       & R & ERP/Cube & \checkmark \\
3D60~\cite{Zioulis2018OmniDepth}           & 2018 & Indoor   & Depth, normal            & $\sim$36K stereo panos  & M & ERP     & \checkmark \\
Structured3D~\cite{Zheng2020Structured3D}  & 2020 & Indoor   & Seg., depth, layout      & 21,835 rooms            & S & ERP     & \checkmark \\
WildPASS~\cite{Yang2021WildPASS}           & 2021 & Outdoor  & Sem. seg. (eval)         & 500 labeled panos       & R & ERP     & \checkmark \\
DensePASS~\cite{Ma2021DensePASS}           & 2021 & Outdoor  & Sem. seg. (UDA)          & 100 labeled panos       & R & ERP     & \checkmark \\
SynPASS~\cite{Zhang2024Trans4PASS}         & 2024 & Outdoor  & Sem. seg. (Syn2Real)     & 9,080 panos             & S & ERP     & \checkmark \\
\midrule
\multicolumn{8}{l}{\textit{Open-World \& Safety}} \\
OPS~\cite{Zheng2024OPS}              & 2024 & In/Outdoor & Open-vocab seg.          & Task protocol$^*$       & -- & ERP    & \checkmark \\
PanOoS~\cite{Duan2025PanOoS}        & 2025 & Outdoor    & OoD seg.                 & DenseOoS + QuadOoS      & R & ERP/PAL & \checkmark \\
\midrule
\multicolumn{8}{l}{\textit{Multimodal \& Reasoning}} \\
Pano-AVQA~\cite{Yun2021PanoAVQA}     & 2021 & In/Outdoor & Audio-visual QA          & 5.4K video clips        & R & ERP     & \checkmark \\
OSR-Bench~\cite{Dongfang2025OSRBench} & 2025 & Indoor    & Spatial reasoning VQA    & 153K QA pairs           & M & ERP     & \checkmark \\
OmniVQA~\cite{Zhang2025OmniVQA360R1}  & 2025 & Indoor     & 360\textdegree{} VQA     & Based on S2D3D          & R & ERP     & \checkmark \\
Dense360~\cite{Dense360_2025} & 2025 & In/Outdoor & Caption., grounding     & 160K panos, 5M captions & R & ERP     & \checkmark \\
PanoEnv~\cite{Pan2026PanoEnv}        & 2026 & In/Outdoor & 3D spatial VQA           & Sim.-grounded QA        & S & ERP     & \checkmark \\
\midrule
\multicolumn{8}{l}{\textit{Video \& Tracking}} \\
360VOT~\cite{Huang2023360VOT}        & 2023 & General    & SOT                      & 120 seq., 113K frames   & R & ERP     & \checkmark \\
360VOTS~\cite{Xu2025360VOTS}         & 2025 & General    & SOT + VOS                & 360VOS: 290 seq.\ / 62 cat.; inherits 360VOT & R & ERP     & \checkmark \\
PanoVOS~\cite{Yan2024PanoVOS}        & 2024 & General    & VOS                      & 150 videos, 19K masks   & R & ERP     & \checkmark \\
QuadTrack~\cite{Luo2025OmniTrack}        & 2025 & Embodied & MOT & 19,200 images       & R & PAL & \checkmark \\
EmboTrack~\cite{Luo2025OmniTrackPlus}    & 2025 & Embodied & MOT & QuadTrack $+$ BipTrack$^\dagger$ & R & PAL &  $^\ddagger$ \\
Leader360V~\cite{Zhang2026Leader360V} & 2026 & Diverse   & Inst. seg. + tracking    & 10K+ videos, 198 cat.   & R & ERP     & \checkmark \\
\bottomrule
\end{tabular}}
\par\smallskip
\begin{minipage}{\textwidth}\scriptsize
$^*$OPS defines a task protocol evaluated on existing benchmarks (WildPASS, Stanford2D3D, Matterport3D).
$^\dagger$EmboTrack is an umbrella benchmark subsuming QuadTrack and BipTrack; QuadTrack is listed separately above.
$^\ddagger$Stated as forthcoming in the source paper at time of writing.
\end{minipage}
\end{table*}

\subsection{Evaluation Protocol Critique}
\label{sec:eval_critique}

Beyond data availability, the quality of evaluation protocols determines whether reported progress reflects genuine spherical understanding.
Current practice is largely inherited from the perspective domain: segmentation is scored with mean Intersection-over-Union (mIoU), depth with RMSE, absolute relative error, and threshold accuracy ($\delta_1$), and tracking with HOTA or success and precision, all computed by treating every ERP pixel as equal and every frame as an independent planar image. These metrics are convenient and comparable across papers, but none is aware of the sphere, and the five gaps below follow from that.

\paragraph{(a) Spherical-area-aware metrics.}
In ERP, pixels near the poles cover disproportionately small spherical areas compared with equatorial pixels, yet standard mIoU and depth RMSE treat all pixels equally.
WS-PSNR~\cite{Sun2017WSPSNR} introduced spherical-area weighting for 360\textdegree{} video quality assessment by multiplying per-pixel errors with a $\cos\varphi$ weight proportional to the solid-angle element.
Pano3D~\cite{Albanis2021Pano3D} explicitly noted that without such weighting, standard depth metrics favor performance in the distorted polar regions.
On the training side, SGAT4PASS~\cite{Li2023SGAT4PASS} introduced a panorama-aware loss weighted by spherical pixel density.
However, we did not find a panoramic segmentation or depth benchmark whose standard, cross-paper reporting protocol \emph{requires} a spherical-area-weighted mIoU or RMSE, so polar regions remain systematically over-represented in nearly all reported results.

\paragraph{(b) Seam-consistency testing.}
The left and right boundaries of an ERP image are an artifact of the cylindrical unwrapping; objects crossing this boundary should receive consistent predictions from both sides. SAP~\cite{Jiang2026SAP} is, to our knowledge, the first work to construct a paper-specific seam diagnostic subset (76 instances drawn from PAV-SOD and HunyuanWorld-1.0) and report significant degradation of vanilla SAM2 there; the OPS~\cite{Zheng2024OPS} authors similarly acknowledge that their architecture does not cover $360^{\circ}$ boundary continuity. Yet no existing benchmark includes a cross-paper seam-consistency metric.

\paragraph{(c) Polar-robustness stratification.}
ERP distorts the polar regions disproportionately, so latitude-band-stratified accuracy is a diagnostic concern distinct from seam continuity. SGAT4PASS~\cite{Li2023SGAT4PASS} proposed Spherical Geometry-Aware (SGA) validation, which tests robustness to 3D rotation perturbations by reporting mIoU mean and variance under pitch and roll disturbances. SGA validation captures rotational invariance but not per-latitude-band accuracy: a model could achieve low SGA variance while still failing systematically at high latitudes. To date, no benchmark includes a latitude-band stratification requirement as a standard reporting protocol.

\paragraph{(d) Cross-projection evaluation.}
While some methods, notably DPPASS~\cite{Zheng2023DPPASS} and 360SFUDA++~\cite{Zheng2024_360SFUDAplus}, exploit multiple projections (ERP, tangent, fixed-FoV) \emph{during training}, no standard panoramic benchmark requires held-out testing on a different projection (e.g., training on ERP and evaluating on cubemap or tangent representations).
This omission makes it impossible to tell whether a model has learned genuine spherical scene understanding or has instead overfit to the spatial statistics and distortion patterns of one specific projection format.

\paragraph{(e) Open-world protocol standardization.}
The emerging open-world panoramic benchmarks lack a unified evaluation protocol.
OPS~\cite{Zheng2024OPS} evaluates with both closed-vocabulary mIoU and open mIoU (using WordNet-based semantic similarity); PanOoS~\cite{Duan2025PanOoS} reports AuPRC and FPR95 for anomaly detection; the panoramic VQA benchmarks (OSR-Bench, OmniVQA, PanoEnv) each define their own question taxonomies, answer formats, and scoring functions.
Without a shared protocol, cross-method comparison is unreliable and progress is hard to measure consistently.
\subsection{Summary}
\label{sec:data_summary}

Three structural limits define the benchmark ecosystem: segmentation data remains small and indoor- and ERP-biased (Stanford2D3D's ${\sim}1.4$K panoramas against ADE20K's ${\sim}25$K perspective images, with outdoor sets smaller still); multimodal and reasoning benchmarks have proliferated since 2025 but each with its own protocol; and the video and tracking family has only just reached large scale with Leader360V, hinting at a path toward data-scale parity. Cutting across all of them, evaluation protocols still do not test spherical-geometry awareness, since none weights metrics by spherical area, tests seam consistency, stratifies accuracy by latitude, or requires cross-projection generalization. These gaps directly motivate the recommendations of Section~\ref{sec:open_problems}.


\section{Open Problems and Future Directions}
\label{sec:open_problems}

Several fundamental problems remain unsolved. They cluster around the tension that runs through this survey, between respecting the geometry of the sphere and reusing the perspective-pretrained models and datasets that drive modern vision, and they will not be closed by scale alone. This section groups six of them into concrete research directions.

\subsection{From Projection Adaptation to Sphere-Native Modeling}
\label{sec:future_sphere_native}

The empty quadrant identified in Section~\ref{sec:unify_compare} (Fig.~\ref{fig:ch3_taxonomy}) corresponds to a concrete trade-off: spectral spherical CNNs~\cite{Cohen2018Spherical} achieve exact SO(3)-equivariance but operate on custom spectral representations that rule out reuse of ImageNet- or CLIP-pretrained backbones, while adapter-based strategies (ERP-RoPE~\cite{Dense360_2025}, the LoRA-tuned SAM2 encoder in OmniSAM~\cite{Zhong2025OmniSAM}, the spatio-modal fusion over a frozen SAM encoder in PanoSAMic~\cite{Chamseddine2026PanoSAMic}, and the M\"obius spatial augmentation in PanDA~\cite{Cao2025PanDA}) preserve pretrained knowledge but only introduce approximate geometric corrections that still degrade near the poles. MTPano~\cite{Zhang2026MTPano} sidesteps the dilemma by projecting panoramas into perspective patches, generating pseudo-labels with off-the-shelf perspective foundation models, and re-projecting them onto the sphere as patch-wise supervision, but this pipeline inherits the geometric blind spots of its perspective teachers. The near-term pragmatic path is parameter-efficient sphere-aware adapters that inject geometric inductive biases into frozen backbones; the longer-term challenge is sphere-native pretraining at foundation scale, comparable in ambition to what DINOv2 or CLIP did for perspective vision.

\subsection{Geometry-Aware Tokenization and Representation at Scale}
\label{sec:future_tokenization}

How panoramic images are tokenized decides what geometric information reaches downstream layers. The GT family already encompasses several axes of variation: positional encoding (ERP-RoPE~\cite{Dense360_2025}, mechanism detailed in Section~\ref{sec:geo_token}), patch embedding (SDPE in SGAT4PASS~\cite{Li2023SGAT4PASS}), spherical-harmonic queries (HUSH~\cite{Lee2025HUSH}, mechanism detailed in Section~\ref{sec:unified}), and learned token mixers with absolute-position injection (MTPano~\cite{Zhang2026MTPano}). SAP~\cite{Jiang2026SAP} takes a different route, converting the panorama into a fixed-trajectory perspective video so that it aligns with SAM2's memory mechanism. Despite this diversity, there is no systematic comparison of these positional-encoding and tokenization variants on a standardized benchmark. Open questions include whether a universal panoramic tokenizer can work across projections, whether token density should be latitude-adaptive, and whether spherical positional encoding can be learned end-to-end rather than hand-designed. Near-term progress requires controlled ablation studies; in the longer term, the goal is tokenizers that embed spherical geometry into the representation itself, not only into the positional encoding.

\subsection{Unified Pretraining Across Dense, Multimodal, and Temporal Tasks}
\label{sec:future_unified}

Current panoramic methods are overwhelmingly task-specific. HUSH~\cite{Lee2025HUSH} and MTPano~\cite{Zhang2026MTPano} unify multiple dense prediction tasks (e.g., depth, surface normals, semantic segmentation, layout) but do not bring in language. Dense360VLM~\cite{Dense360_2025} covers vision-language captioning and grounding on panoramas but lacks dense prediction heads. OmniTrack~\cite{Luo2025OmniTrack} and its extension OmniTrack++~\cite{Luo2025OmniTrackPlus} handle panoramic multi-object tracking but operate independently of semantic understanding or language grounding. In perspective vision, unified models such as GPT-4o~\cite{OpenAI2024GPT4o} jointly process text, images, audio, and video in a single architecture, and the broader push toward unified multimodal understanding and generation~\cite{zhao2025unified} is moving fast. The panoramic counterpart does not yet exist. In the near term, multi-task panoramic pretraining on large-scale indoor datasets, combining depth, segmentation, and spatial VQA, would consolidate currently isolated advances. Further out, the field needs panoramic world models that jointly reason about geometry, semantics, language, and temporal dynamics within a single framework.

\subsection{Data Ecosystem: Annotation, Synthetic Data, and Evaluation Reform}
\label{sec:future_data}

Three connected data problems slow progress. The first is annotation scarcity: a widely used real-world indoor panoramic segmentation dataset, Stanford2D3D~\cite{Armeni2017Stanford2D3D}, contains only about 1,400 images with 13 categories, while perspective benchmarks such as ADE20K~\cite{Zhou2017ADE20K} exceed 25K images with 150 categories. Leader360V's A3360V pipeline~\cite{Zhang2026Leader360V}, which coordinates 2D segmentors and LLMs for foundation-model-assisted auto-labeling, and SAP's procedural InfiniGen-based synthesis~\cite{Jiang2026SAP} show that automated annotation can scale dataset construction, but neither has been validated for cross-domain generalization. The second is the synthetic-to-real gap: Structured3D~\cite{Zheng2020Structured3D} and SynPASS~\cite{Zhang2024Trans4PASS} provide large-scale training data, yet models trained on them consistently underperform on real-world benchmarks. The third is evaluation: the five protocol gaps identified in Section~\ref{sec:eval_critique} remain unadopted in standard reporting. Adopting spherically-weighted mIoU together with seam and polar diagnostics would already improve rigor; over a longer horizon, a panoramic evaluation suite analogous to COCO should standardize metrics, splits, and protocols.

\subsection{Cross-Projection Generalization}
\label{sec:future_cross_proj}

As Section~\ref{sec:eval_critique} noted, no benchmark requires held-out testing on a projection other than the training one, a gap the PANORAMA survey~\cite{Zheng2025PANORAMA} also flags. Real-world deployment demands flexibility: VR headsets typically render in cubemap, surveillance systems use fisheye, and robotics platforms employ diverse formats. A method that performs well only on ERP has likely learned ERP-specific artifacts (horizontal stretching, polar compression) rather than genuine spherical understanding. This problem is specific to panoramic vision; perspective models face no analogous challenge because a single projection dominates their domain. Cross-projection evaluation should become a standard benchmark protocol, in the same spirit as the SGA validation proposed by SGAT4PASS~\cite{Li2023SGAT4PASS} for rotation robustness. In the longer term, projection-agnostic architectures that process the underlying spherical signal directly, regardless of how it is stored or transmitted, would resolve the issue at its root.

\subsection{Toward General-Purpose Panoramic Intelligence}
\label{sec:future_general}

Panoramic perception, multimodal reasoning, and temporal understanding currently exist as disconnected research threads. A general-purpose panoramic AI system would need at minimum four capabilities working together: (a) sphere-native visual encoding that respects spherical geometry, (b) 360\textdegree{} spatial reasoning grounded in natural language, (c) temporal persistence and object tracking across panoramic video, and (d) embodied interaction for navigation and manipulation. No existing system integrates all four. PanoEnv~\cite{Pan2026PanoEnv} and 360-R1~\cite{Zhang2025OmniVQA360R1} advance panoramic spatial reasoning through reinforcement learning on VQA tasks; OmniSAM~\cite{Zhong2025OmniSAM} and SAP~\cite{Jiang2026SAP} push foundation-model perception toward 360\textdegree{} imagery; PanoAffordanceNet~\cite{zhu2026PanoAffordance} takes a first step toward scene-level affordance grounding for embodied intelligence. What is still missing is the integration of temporal and embodied panoramic intelligence with these perceptual and linguistic components. Over a three-to-five-year horizon, the community should pursue end-to-end panoramic agents that perceive, reason, and act in 360\textdegree{}, trained on diverse panoramic data with unified geometric, linguistic, and temporal supervision.

 
\section{Conclusion}
\label{sec:conclusion}
 
This survey has organized panoramic scene understanding along two orthogonal axes, architectural design and training paradigm, and traced a trajectory of deepening geometric commitment from projection-based adaptation, through distortion-aware engineering, to sphere-native modeling, with geometry-aware tokenization emerging as a distinct foundation-model-era family. A single structural tension runs through this trajectory: exact spherical equivariance and full reuse of perspective-pretrained foundation-model weights remain mutually exclusive. The field has resolved it not by adopting the theoretically strongest sphere-native operators, which attain exact rotation equivariance but cannot reuse pretrained weights and have therefore never scaled, but by settling on a compatibility-preserving middle ground that couples moderate geometric awareness with large pretrained models; every one of the strongest 2024--2026 advances we surveyed operates in this regime.
 
Beyond the architectural trade-off, panoramic scene understanding has expanded rapidly but unevenly along the task axis. Dense prediction has matured from closed-set segmentation and depth estimation to unified multi-task understanding via spherical harmonics, foundation-model-assisted open-world perception, and, most recently, vision-language reasoning grounded in 360\textdegree{} spatial context; within it, foundation-model adaptation has advanced depth fastest while leaving layout and surface-normal estimation largely untouched. Dynamic panoramic perception, by contrast, remains at an earlier stage: the dominant pattern is spatial sphere-awareness combined with temporal planarity, and no method yet formulates temporal association natively on the sphere. Cutting across all tasks, no panoramic foundation model has yet been pretrained directly on spherical data; every method surveyed initializes from perspective-pretrained weights.
 
Five gaps in evaluation infrastructure, namely the absence of spherical-area-weighted metrics, seam-consistency testing, polar-robustness stratification, cross-projection generalization, and standardized open-world protocols, mean that reported progress does not fully reflect genuine spherical understanding. Addressing these gaps is a prerequisite for reliable measurement of future advances.
 
The research roadmap points toward one overarching goal: building general-purpose panoramic intelligence that perceives, reasons, and acts in the full visual sphere. Getting there requires co-designing sphere-native representations with foundation-model-scale training, developing unified pretraining across geometric, linguistic, and temporal tasks, and establishing an evaluation ecosystem that tests what panoramic models actually need to do, not only what current benchmarks happen to measure.

\bmhead{Acknowledgements}

This work was supported by the Xi'an Jiaotong-Liverpool University Postgraduate
Research Scholarship under Grant No. FOS2210JJ03.

\bibliography{sn-bibliography}

\end{document}